\pdfoutput=1
\documentclass{article}

\usepackage{arxiv}

\usepackage[utf8]{inputenc} 
\usepackage[T1]{fontenc}    
\usepackage{hyperref}       
\usepackage{url}            
\usepackage{booktabs}       
\usepackage{amsfonts}       
\usepackage{nicefrac}       
\usepackage{microtype}      
\usepackage{lipsum}		
\usepackage{graphicx}
\usepackage{natbib}
\usepackage{doi}

\usepackage[utf8]{inputenc} 
\usepackage[T1]{fontenc}    
\usepackage{hyperref}       
\usepackage{url}            
\usepackage{booktabs}       
\usepackage{amsfonts}       
\usepackage{nicefrac}       
\usepackage{microtype}      
\usepackage{lipsum}
\usepackage{graphicx}
\usepackage{amsmath,amssymb,amsfonts}
\usepackage{graphicx}
\usepackage{textcomp}
\usepackage{xcolor}
\usepackage{mathtools, cuted}
\usepackage{multirow}
\usepackage{color,soul}
\usepackage{xspace}
\usepackage{subfig}
\usepackage{array}
\usepackage{algorithm}
\usepackage{algpseudocode}
\usepackage{hyperref}
\usepackage[toc,page]{appendix}
\usepackage{adjustbox}

\title{Optical tracking in team sports}

\author{
 Pegah Rahimian \\
  Budapest University of Technology\\ and Economics\\
  Budapest, Hungary \\
  \texttt{pegah.rahimian@tmit.bme.hu} \\
  \And
 Laszlo Toka \\
  MTA-BME Information Systems\\ Research Group\\
  Budapest, Hungary\\
  \texttt{toka.laszlo@vik.bme.hu} \\
}

\begin{document}
\maketitle

\begin{abstract}
	Sports analysis has gained paramount importance for coaches, scouts, and fans. Recently, computer vision researchers have taken on the challenge of collecting the necessary data by proposing several methods of automatic player and ball tracking. Building on the gathered tracking data, data miners are able to perform quantitative analysis on the performance of players and teams. With this survey, our goal is to provide a basic understanding for quantitative data analysts about the process of creating the input data and the characteristics thereof. Thus, we summarize the recent methods of optical tracking by providing a comprehensive taxonomy of conventional and deep learning methods, separately. Moreover, we discuss the preprocessing steps of tracking, the most common challenges in this domain, and the application of tracking data to sports teams. Finally, we compare the methods by their cost and limitations, and conclude the work by highlighting potential future research directions.
\end{abstract}

\keywords{sports analytics \and player recognition  \and  player tracking \and  optical tracking \and image processing \and deep learning   }

\section{Introduction}

The success in every team sport significantly depends on the analysis of the semantics.
Most team sports, such as football, basketball, and ice hockey, involve very complex interactions between players. 
Researchers and data analysts propose various methods for modeling these interactions. For this aim, they need to follow the movements of players and the ball from the video. However, this task is strenuous due to the large speed of players and of the ball in the playfield, and tracking usually fails in the cases of overlaps, poor light conditions, and low quality of the videos. During the past decades, computer vision researchers developed several optical tracking algorithms by analyzing video image pixels and by extracting the features of the objects of interest, such as players and the ball. On this data, the movement, action, intention, and gesture of the players can be analyzed. 

The most common analysis is performed over player and ball tracking data, also known as trajectory data. 
The distilled knowledge can help coaches and scouts in several aspects, such as game strategy and tactics, goal analysis, pass and shot prediction, referee decisions, player evaluation, and talent identification. 
In order to automatize the end-to-end analytics procedure, the tracking methods require visual data (video frames) as the data source and produce tracking data (player and ball trajectories) for further data mining. 
The proposed methods majorly contribute to effectively evaluate the performance at individual and team levels in team sports. 
E.g., at the individual level, the characteristic style of a player, while at the team level, the combination of all players' trajectories can be evaluated. 

The work in this paper is motivated by the following observations. First, researchers in sports analytics are continuously searching for the most accurate, but a cost-effective method for the player and ball tracking. The above-mentioned goals of tracking prove the importance of opting for an accurate method for extracting player and ball trajectories in sports analytics. Second, player and ball tracking are one of the broadest areas for research in sports analytics. In the literature, there are many published works without proper classification. Recently, the automatic feature extraction capability of deep learning in computer vision encourages sports analysts to experiment with neural networks for player and ball tracking tasks. Thus, a wider range of tracking options are available to the researchers and this survey helps them to choose their suitable method depending on the task at hand. Furthermore, understanding all these methods requires deep knowledge of computer vision for quantitative analysts in sports, which is not realistic. Therefore, in this paper, we have the following goals: to provide a robust classification of methods for the two tasks of detection and tracking and to give insights about the applied computer vision techniques of extracting trajectories to the quantitative analysts in sports.

Several papers made attempts to present the myriad of state-of-the-art object tracking algorithms. A broad description of object tracking methods was given in \citet{yilmaz2006}, and a more recent one in \citet{RasoolReddy2015}. Moreover, \citet{Dhenuka2018} presented a survey on Multiple Object Tracking (MOT) methods, while a survey for solving occlusion problems was published in \citet{Lee2014}. The first survey on the application of deep learning models in MOT is presented in \citet{Gioele2019}. All these surveys cover the description of tracking methods of generic objects, such as humans or vehicles. It was in \citet{Manafifard2017} where the authors summarized the state-of-the-art player tracking methods focusing on soccer videos. Although, these surveys show the following shortcomings. Most of these papers are not dedicated to team sports and survey all kinds of object tracking algorithms. On the other side, the sport dedicated survey like \citet{Manafifard2017}, is too technical, suitable only for computer vision analysts, and dedicated to tracking.

This survey contributes to the state-of-the-art player and ball tracking methods as follows. First, the methods in detection and tracking tasks are classified separately. Second, this paper is not only listing the methods but also gives an insight about the computer vision techniques to the quantitative analysts in sports, who need the extracted trajectories for their quantitative models. Third, the application of deep learning in team sports is surveyed for the first time in the literature. Fourth, we provide a cost analysis of the methods according to their computational and infrastructure requirements.       

This paper is organized as follows. In Section~\ref{camera} we explain our paper collection process and the camera setup requirements of the published works. We list the methods for the player and ball detection in Section~\ref{sec:det}, and the player and ball tracking in Section~\ref{sec:tra}. We evaluate the categorized techniques in terms of their applied theoretical methods and analyze their cost in Section~\ref{sec:eval}, and finally, we conclude the work in Section~\ref{sec:ind}.

\section{Eligibility and data collection}
\label{camera}

This survey is conducted to help quantitative sports analysts choose the best method to create their own tracking data from sports videos. For this task, the eligible papers are collected from Science Direct, Google Scholar, Scopus databases, and ACM, IEEE, Springer digital libraries using the following keywords for filtering papers and minimizing bias: ``Sports analytics'', ``soccer'', ``player tracking'', ``ball tracking'', ``player detection'', ``ball detection'', ``deep learning for tracking'', ``fixed camera'', ``moving camera'', ``broadcast sports video''. In the first round of collection, 125 papers have been identified and we carefully inspected their contributions in terms of 1) detection or tracking, 2) camera setup, and 3) deep learning-based or traditional methodologies. In order to make the best structure of this survey, we excluded the papers in which tracking was not the main focus. An example is a method called DeepQB in American football proposed by \citet{Bryan2019}. This paper proposes a deep learning approach applied to player tracking data to evaluate quarterback decisions, which is clearly not a direct contribution in player tracking methods. As a result of filtering those papers and focusing on player or ball detection and tracking, 50 papers were eligible for this survey. Furthermore, we also classified eligible papers according to their camera setup as follows.  


One of the most important criteria for the evaluation of the methods in this work is the required camera setup. Depending on the camera setup, the frame extraction methods are different. Several studies in sports video analytics are limited to a single fixed camera. In these methods preprocessing steps are simpler and faster, as they do not require time and location synchronization. However, as they need to cover the whole playfield, the frames are mostly blurry and difficult to use for detection \citet{Needham2001, Gonzalo2012, SabirinHiroshi2015, Adria2019}. An alternative setting to improve resolution and accuracy is to use multiple fixed cameras. In these videos occlusion problems can be handled easily, as the occluded player or ball in one frame can be recognized with the frame captured by another camera from other angles \citet{Jinchang2008, Jinchan2009, Lan2008, Yazdi2018}. Another option is to use multiple moving cameras, which makes the video processing more complex, but it provides more flexibility in the analysis. These types of video require significant synchronization effort, but finally, they produce longer trajectories, as the cameras try to follow ball controllers \citet{Ming2004,Neus2010,Dipen2014,hosseinAlavi2017}. In this paper, we classify each of the citetd papers according to their required video inputs in terms of the cameras being fixed or moving, and of their cardinality in the arena. 

\section{Player and ball detection}
\label{sec:det}

\begin{figure}[h]
        \center{\includegraphics[width=.7\textwidth]{ 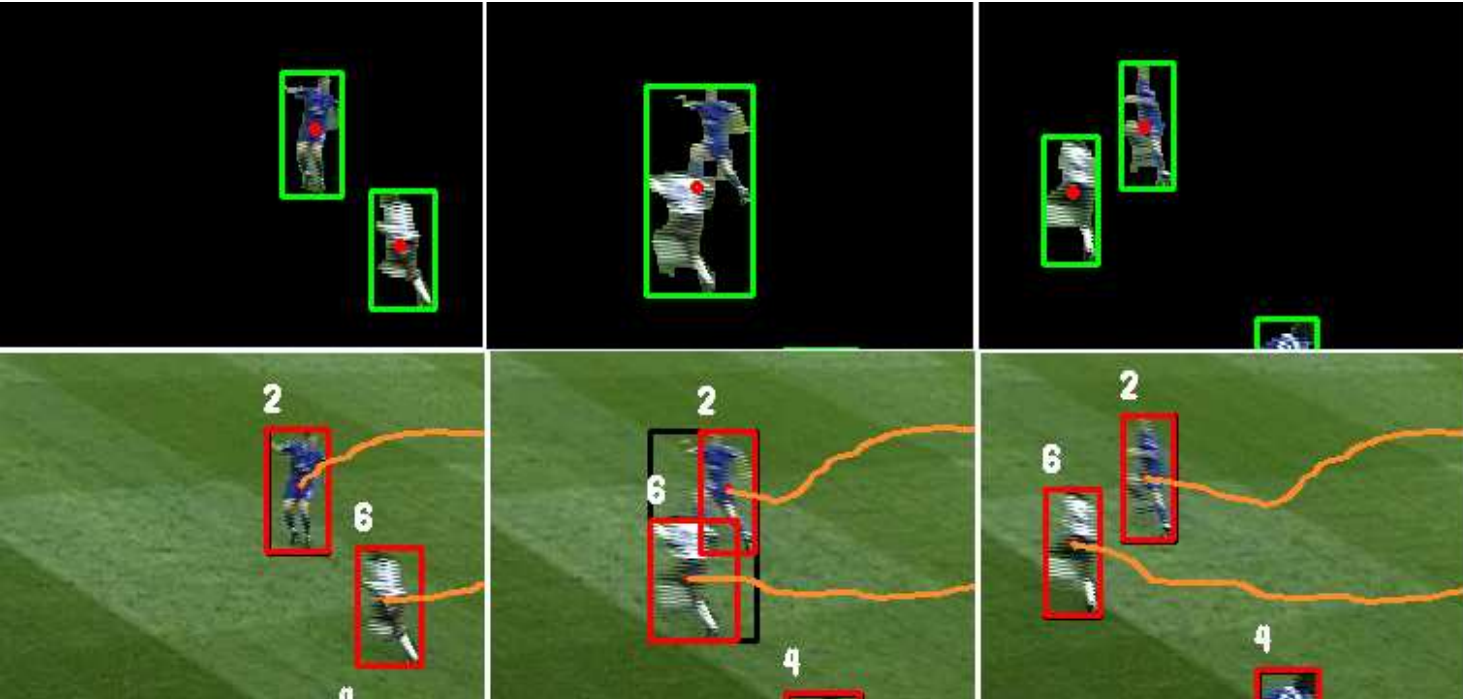}}
        \captionsetup{justification=centering}
        \caption{\label{bb}Player detection (top) and tracking (bottom) results from \citet{Ming2004}}
      \end{figure}

Tracking data, i.e., the exact location of the players and the ball on the field at each moment of the match, is the most important data for a quantitative model developer. Player and ball detection methods are computer vision techniques that allow the analyst to identify and locate players and the ball in a frame of a sports video. Detection methods provide the input to tracking, which would be a simple task if all players and the ball were totally visible in each frame and there were no occlusion. However, in real-world videos, most frames are blurry and continuous tracking fails due to e.g., occlusion, poor light, or posture changes. Therefore, the detection task should be combined with an appropriate tracking method to accurately track the players and the ball (See Figure~\ref{bb}).

\begin{figure}[h]%
    \centering
    \subfloat[\centering Traditional]{{\includegraphics[width=8.5cm]{ 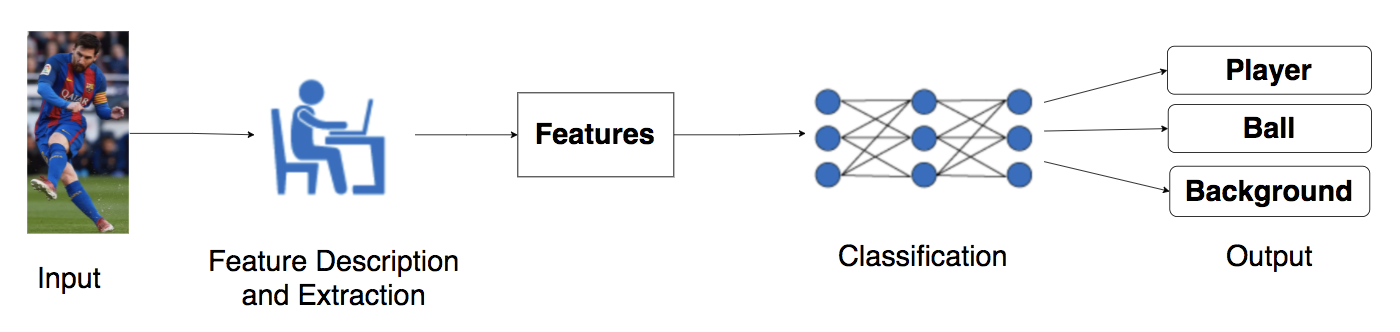} }}%
    \qquad
    \subfloat[\centering Deep Learning]{{\includegraphics[width=7cm]{ 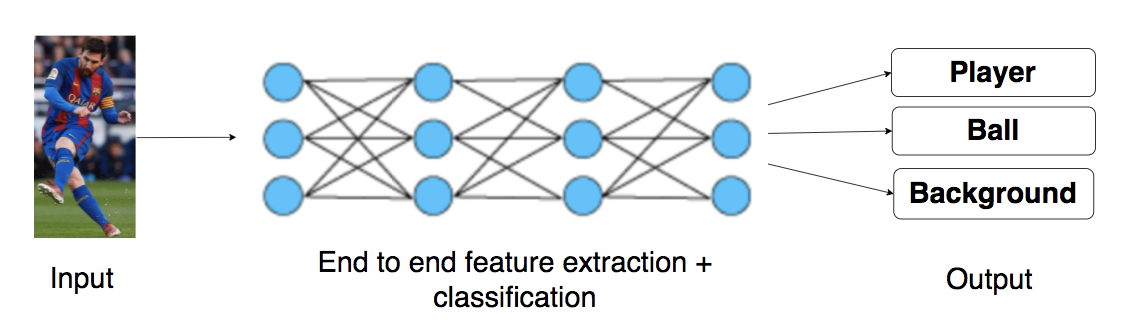} }}%
    \captionsetup{justification=centering}
    \caption{Player and ball detection workflow}%
    \label{wf}%
\end{figure}

In this section, we focus on detection methods that aim to find the bounding box of the players and the ball, and to localize the different detection features inside each bounding box. Bounding boxes are imaginary boxes around players and the ball (see Figure~\ref{bb}) that are used to separate each player and ball from other objects in a video frame. We classify detection methods into the categories of traditional and deep learning-based methods. As Figure~\ref{wf} shows, while in the traditional methods the features of the input objects need to be described and extracted by the analyzer and depend on the detection algorithms, a deep learning method performs this process automatically through the layers of a neural network. Therefore, data quality, computational power, domain expertise, training time, and required accuracy specify the selection of the suitable choice of method to apply. We briefly describe each group of methods separately, and give a summary of published research papers, along with their important attributes, in
Table~\ref{table:detection}. 

\begin{table}[]
\caption{Review of playfield and player detection methods} 
\centering
\begin{adjustbox}{width=1.\textwidth,center=\textwidth}
\small
\label{table:detection}
\begin{tabular}{|>{\centering\arraybackslash}m{1.5cm}|>{\centering\arraybackslash}m{2cm}|>{\centering\arraybackslash}m{4cm}|>{\centering\arraybackslash}m{4cm}|>{\centering\arraybackslash}m{4cm}|}
 \hline
 \textbf{Reference} & \textbf{Playfield detection} & \textbf{Player detection} &  Team sport \& camera type & \textbf{Evaluation}  
 \\ 
  \hline
        \citet{Slawomir2010} & Hough transform for court line detection & HOG descriptor  &  Football video broadcast  &  performs well in SD and HD test sequences, different light condition, and various positions of the cameras ; 78\% precision  
        \\ 
        \hline
        \citet{Evan2015} & Hough transform  & Pedestrian detection with HOG \& color-based detection  & Basketball video broadcast & miss rate: 70\% for pedestrian detection   
        \\ 
        \hline
        \citet{Ming2009} & Peak value of RGB & Adaptive GMM &  Football video with single moving camera &  Powerful segmentation result, but only in the absence of shadows 
        \\ 
        \hline
        \citet{Mazzeo2008} & Background subtraction  & Moving object segmentation by calculating energy information for each point &  Football video with single stationary camera & Copes with light changes by proposing pixel energy evaluation 
        \\ 
        \hline
        \citet{Direkoglu2018} & Binary edge detection of court line with Canny edge detector & Using shape information of an object by solving heat diffusion equation & Hockey video with single stationary camera & Highly accurate method between 75\% to 98\%, but  computationally less efficient in time required for detection  
        \\ 
        \hline
        \citet{Naushad2012,Upendra2015} & RGB color extraction if G>R>B & Sobel gradient algorithm &  Football video broadcast & Accurately detects the ball when it is attached to the lines; but in crowded places, it fails to detect the player 
        \\ 
        \hline
        \citet{Branko2015} & - & Face recognition with adaboost & Basketball video with single moving camera &Detection accuracy: 70\%\\
        \hline
        \citet{Guangyu2006} & GMM & SVM for player classification &  Soccer, hockey, American football video broadcast & Detection accuracy: 91\% 
        \\
        \hline
        \citet{Chengjun2018} & Background subtraction & One-class SVM & Football video broadcast &  
        Proposes automatic labeling of training dataset that significantly reduces cost and training time 
        \\ 
        \hline
        \citet{Sebastian2015} & - & CNN for number recognition & Football video broadcast & Number level accuracy: 83\%   
        \\ 
        \hline
        \citet{Gen2018} & - & CNN for classification \& Spatial Transformer Network for localization of jersey numbers & Live football video with single moving camera & Number level accuracy: 87\% \& digit level accuracy: 92\%   
        \\ 
        \hline
        \citet{Hengyue2019} & Region Proposal Network & R-CNN for digit localization and classification &  Footbal video with single pan-tilt zooming camera & Number level accuracy: 92\% \& digit level accuracy: 94\%   
        \\ 
        
        \hline

\end{tabular}
\end{adjustbox}
\end{table} 

\subsection{Traditional methods for detection}
In the traditional methods of detection, the features of players, ball, and playfield must be precisely described and extracted by the analyzer. In this section, we classify the methods according to their description of the features, and their extraction types.

\subsubsection{Histogram of Oriented Gradients}
\begin{figure}[]
        \center{\includegraphics[width=8.5cm]{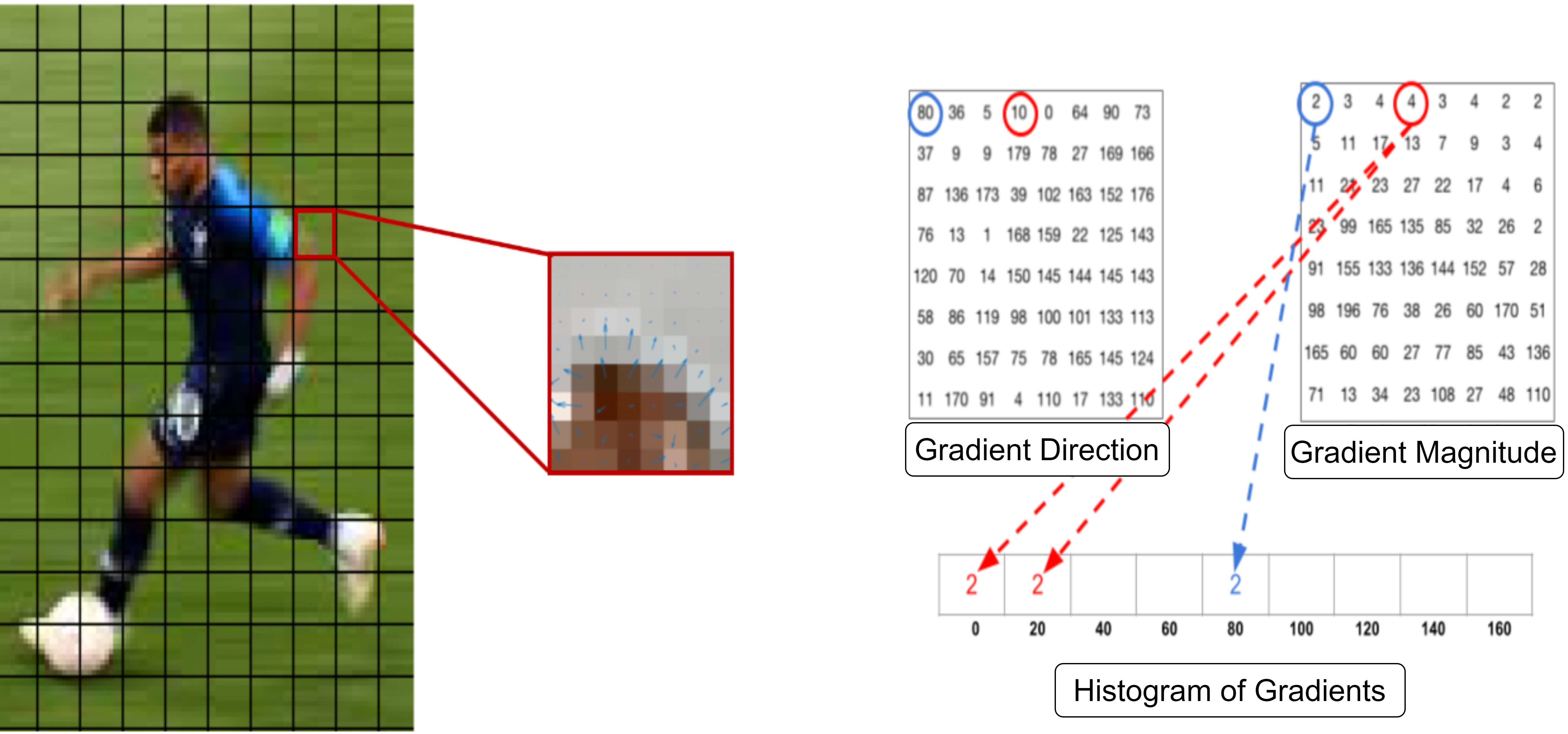}}
        \captionsetup{justification=centering}
        \caption{\label{fighog}Calculating Histogram of Gradients}
      \end{figure}

Histogram of Oriented Gradients \uppercase{(hog)} is a feature descriptor and is essentially used to detect multiple objects in an image by building histograms of pixel gradients through different parts of the image. HOG considers these oriented gradients as features. An example of a calculating histogram of gradients is illustrated in Figure~\ref{fighog}. As the first step, the frame is divided into $8\times8$ cells. For each cell, the gradient magnitude (arrows' length) and gradient direction (arrows' direction) will be identified. Consequently, the histogram containing 9 bins corresponding to angles 0, 20, 40, \dots, 160 is calculated. This feature vector can be used to classify objects into different classes, e.g., player, background, and ball. This method is used by \citet{Slawomir2010} and \citet{Evan2015}.

In these methods, the court lines can be detected with Hough transform, another feature extraction technique that searches for the presence of straight lines in an image. This algorithm fits a set of line segments to a set of image pixels.

\subsubsection{Background modelling}
Background modeling is another method for detecting players and the ball, and is a complex task as the background in sports videos frequently changes due to camera movement, shadows of players, etc. Most of the methods in the background modeling domain consider image pixel values as the features of the input objects. In the domains of player and ball detection, the following two methods are proposed by researchers for background modeling: Gaussian Mixture Model (GMM) and Pixel energy evaluation..

\textbf{Gaussian Mixture Model (GMM): }
GMM is proposed by \citet{Ming2009} where playfield detection is performed first by taking the peak values of RGB histograms through the frames. This is because they assume the playfield is the largest area in the frames. Then each of these extracted background pixels is modelled by $k$ Gaussian distributions; different Gaussians are for different colors. Thus, the probability of a pixel having value $X_t$ can be calculated as:
\begin{equation} \label{eq1}
P(X_t)= \sum_{i=1}^{k} \omega_i \eta (X_t)
\end{equation}
where $\omega_i$ is the weight for the $i^{th}$ component (all summing to 1), and $\eta(X_t)$ is a normal distribution density function. Based on these probabilities and by setting arbitrary thresholds on the value of the pixels, the background pixels can be subtracted and the players or the ball will be detected.
This algorithm cannot recognize players in shadows.\\

\textbf{Pixel energy evaluation}: Another background model is proposed by \citet{Mazzeo2008}. In this method, the energy information of each point is analyzed in a small window: first, the information, i.e, mean and standard deviation, of the pixels at each frame is calculated. Then, by subtracting the information of the first image of the window and each subsequent image, the energy information of each point can be identified. Consequently, the slower energy points (static ones) represent the background, and higher energy points (moving ones) represent the players or the ball.

\subsubsection{Edge detection}
\begin{figure}[ht]%
    \centering
    \subfloat[]{{\includegraphics[height=2cm]{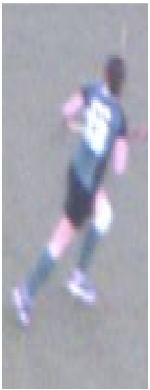} }}%
    \qquad
    \subfloat[]{{\includegraphics[height=2cm]{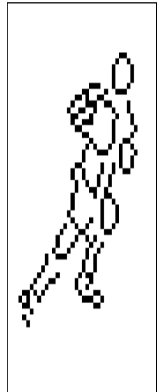} }}%
    \qquad
    \subfloat[]{{\includegraphics[height=2cm]{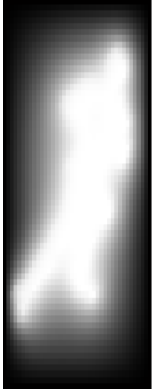} }}%
    \qquad
    \subfloat[]{{\includegraphics[height=2cm]{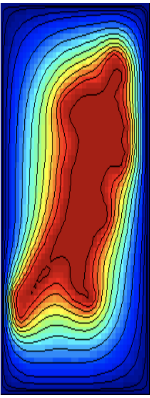} }}%
    \qquad
    \subfloat[]{{\includegraphics[height=2cm]{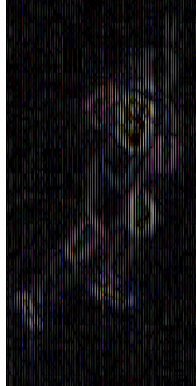} }}%
    \qquad
    \subfloat[]{{\includegraphics[height=2cm]{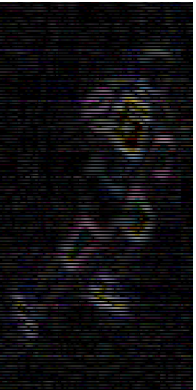} }}%
    \caption{Edge detection methods: (a) original frame, (b) binary edges with Canny method, (c) shape information image \citet{Direkoglu2018}, (d) colored shape information image \citet{Direkoglu2018}, (e) horizontal Sobel operator, (f) vertical Sobel operator}%
    \label{edge}%
\end{figure}

Edge detection is a method for detecting the boundaries of objects within frames as the features. This method works by detecting discontinuities in brightness. The researchers who choose this method for players and ball detection, mostly utilize the following 2 operators: Canny edge detector, and Soble filtering. Figure~\ref{edge} demonstrates the edge detection methods on a sample frame of a player.

\textbf{Canny edge detection:} is a popular method in \texttt{OpenCV} for binary edge detection. (Figure~\ref{edge}(b)).
\citet{Direkoglu2018} proposed using the Canny edge detection method for extracting image data and features. However, there might be missing or disconnected edges, and it does not provide shape information of the players and the ball. 
Thus, given a set of binary edges, they solve a particular heat equation to generate a shape information image (Figure~\ref{edge}(c, d)). In mathematics, the heat equation is a partial differential equation that demonstrates the evolution of a quantity like heat (here heat is considered as binary edges) over time. The solution of this equation is filling the inside object shape. This information image removes the appearance variation of the object, e.g., color or texture, while preserving the information of the shape. The result is the unique shape information for each player, which can be used for identification. This method works only for videos recorded with fixed cameras.

\textbf{Sobel filtering:}
In the method by \citet{Naushad2012} and \citet{Upendra2015}, the Sobel gradient algorithm is used to detect horizontal and vertical edges (Figure~\ref{edge}(e, f)). The gradient is the vector with the components of (x,y) and the direction is calculated as $\tan^{-1} ({\Delta}y / {\Delta}x)$.
Due to the similar color of the ball and the court lines, if the Sobel gradient algorithm is applied for background elimination instead of color segmentation, overlapping of the ball and court lines will not be a problem. However, general overlapping problems, e.g., player occlusion, cannot be handled with this method.

\subsubsection {Supervised learning}

\begin{figure}[h]
        \center{\includegraphics[scale=0.5]{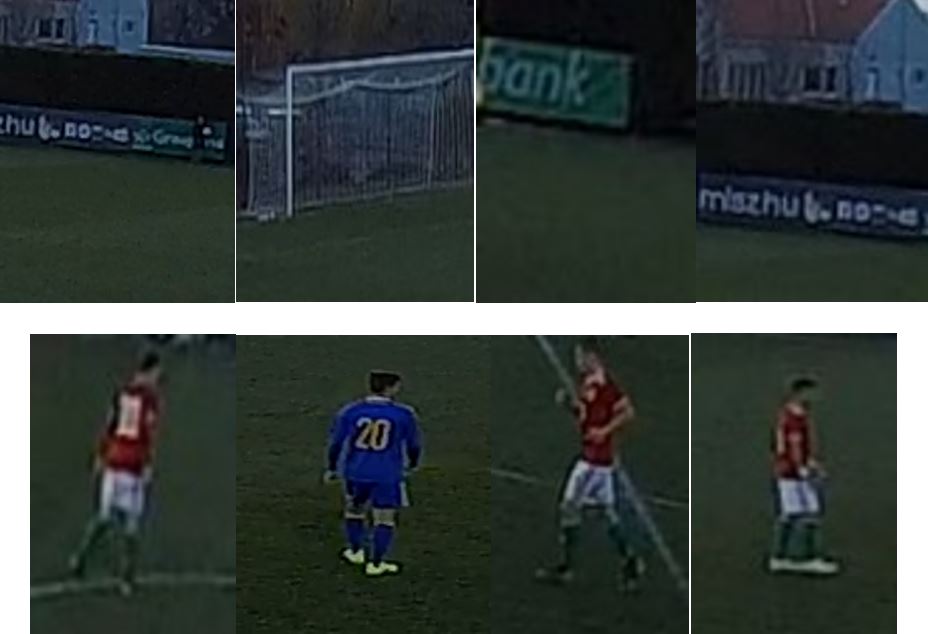}}
        \captionsetup{justification=centering}
        \caption{\label{posneg} Positive (bottom) and negative (top) samples for training classifier}
      \end{figure}

In many proposed methods a robust classifier is trained to distinguish positive samples, i.e., players and/or ball, and negative samples, i.e., other objects or parts of the playfield. Any classification method, such as Support Vector Machine or Adaboost algorithms, can be trained for accurate detection of the players. Some examples of positive and negative sample frames are given in Figure~\ref{posneg}.

\textbf{Support Vector Machine:} Several related works state that the advantages of \uppercase{svm} compared to other classifiers include better prediction, unique optimal solution, fewer parameters, and lower complexity.
In the method of \citet{Guangyu2006}, the playfield is subtracted with a GMM. The results of background subtraction are thousands of objects, which SVM can help to classify into player and not player objects. However, in this method, the training dataset is manually labelled, which is time-consuming. In order to solve this problem, \citet{Chengjun2018} proposed fuzzy decision making for automatic labelling of the training dataset.


\textbf{Adaboost algorithm:} Adaboost, short for Adaptive Boosting is used to make a strong classifier by a linear combination of many weak classifiers to improve detection accuracy. The main idea is to set the weights of the weak classifiers and to train on the data sample in each iteration until it can accurately classify the unknown objects. \citet{Branko2015} used this algorithm for basketball players' face and body parts recognition. Although, they concluded that Adaboost is not accurate enough for object detection in sports events. Furthermore, \citet{Antoine2007} showed that deep learning methods outperform the Adaboost algorithm for player detection.

\subsection{{Deep learning methods for detection}}
\begin{figure}[h]
        \center{\includegraphics[ width=10cm]{ 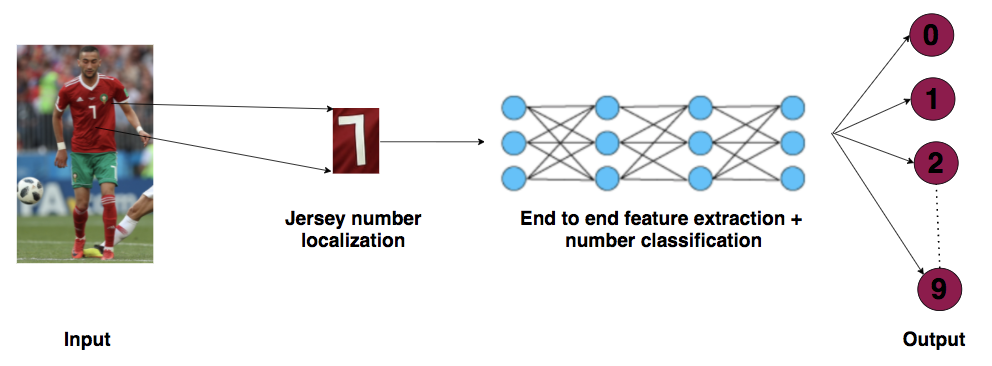}}
        \captionsetup{justification=centering}
        \caption{\label{nn} Neural network architecture for digit localization and detection}
      \end{figure}

In the task of player detection, researchers usually use deep learning to recognize and localize jersey numbers. Most of the works in this area use a Convolutional Neural Network (CNN) which is a deep learning model. The general architecture of CNN for digit recognition is illustrated in Figure~\ref{nn}. As the first step, players' bounding boxes should be detected. Then digits inside each bounding box should be accurately localized. These localized digits will be the input of CNN. Several convolution layers in CNN will assign importance to various features of the digits. Consequently, the neurons in the last layer will classify the digits from 0 to 9 classes. In this area, different works propose the following methods for improving the performance of detection: 1) how to localize digits inside each frame, 2) how to recognize multiple digits, 3) how to automatically label the training dataset, i.e, which benchmark dataset to use.

The first CNN-based approach for automatically recognizing jersey numbers from soccer videos was proposed by \citet{Sebastian2015}.
However, this method cannot recognize numbers in case of perspective distortion of the camera. To solve this problem, \citet{Gen2018} used a Spatial Transformer Network (STN) to localize jersey numbers more precisely. STN helps to crop and normalize the appropriate region of the numbers and improves the performance of classification. Another digit localization technique is Region Proposal Network (RPN), which is a convolutional network that has a classifier and a regressor, and is trained end-to-end to generate high-quality region proposals for digits. RPN is used by \citet{Hengyue2019} for classification and bounding-box regression over the background, person, and digits.
While these methods can be more accurate than some traditional methods for player detection and they eliminate the necessities of manual feature description and extraction, they are also more expensive due to more computation and training time. Most of these methods require special versions of GPUs to be applied. Moreover, training and testing CNNs might be more time-consuming than running traditional methods.

\section{Player tracking}
\label{sec:tra}

Detection methods calculate the location of each player and the ball at each frame of the videos. There are always some frames for which the detection fails due to the blurriness of the frame, poor light conditions, occlusions, etc.  In these cases, the detection methods cannot provide the location of the same player and ball in consecutive frames to construct continuous trajectories. Therefore, a player tracking method is needed to associate the partial trajectories, and to provide long tracking information of each of the players and the ball (see Figure~\ref{bb}). Player tracking involves the design of a tracker that can robustly match each observation to the trajectory of a specific player. This tracker can be designed for a single object or for multiple objects. The biggest challenge in tracking is the overlapping of players, namely the occlusion. Several studies suggested solutions for making a unique, continuous trajectory for each player by solving the occlusion problem. Those methods mostly follow filtering and data association. However, each method follows a different description for interest points (features) for filtering, and data association depends on the custom definition of probabilistic distributions. In this section, we survey the tracking methods classified by whether they are based on traditional or deep learning models. 

\subsection{Traditional methods for tracking}
Same as the previously mentioned traditional detection models, the traditional tracking algorithms also require manual extraction and description of the player and ball features. The main categories of tracking methods in the literature of sports analytics are the following: point tracking, contour tracking, silhouette tracking, graph-based tracking, and data association methods.
\subsubsection{Point tracking}
\begin{figure}[h]
        \center{\includegraphics[scale=0.5]{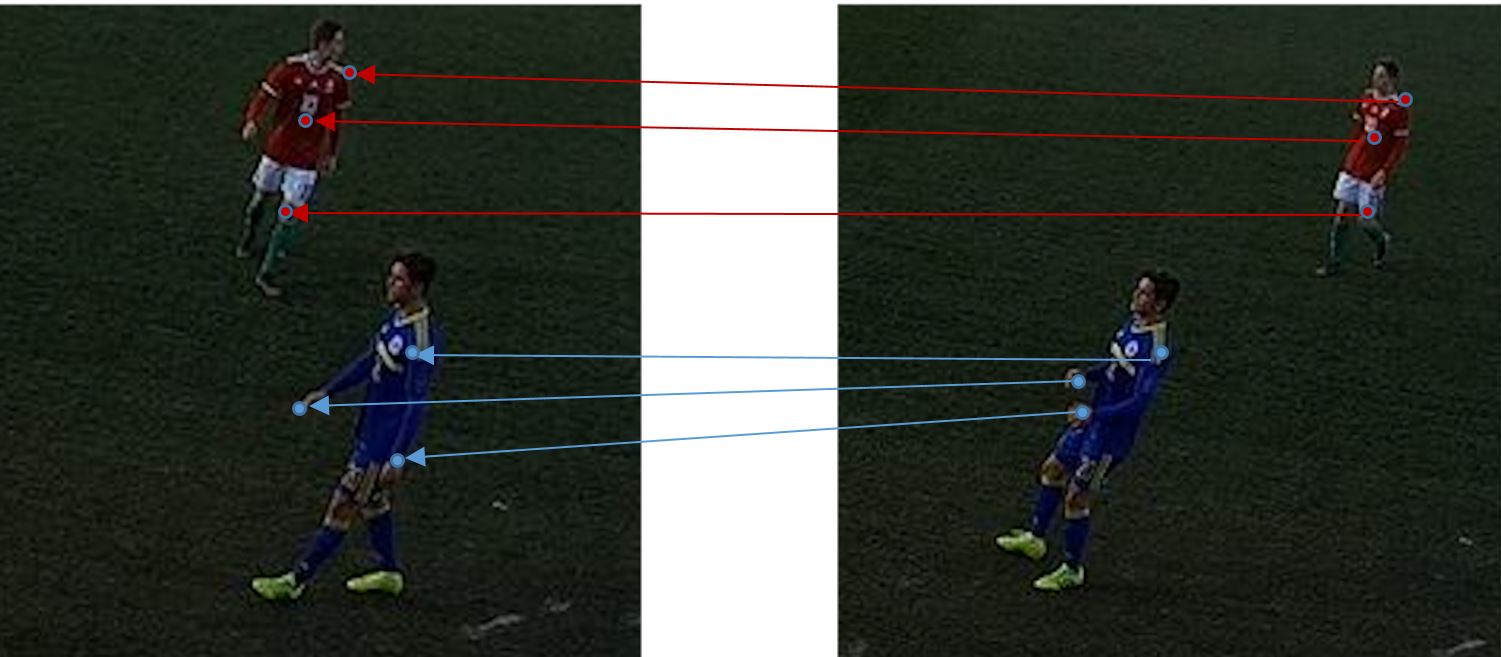}}
        \captionsetup{justification=centering}
        \caption{\label{Point} Point tracking}
      \end{figure}

The methods using point tracking mostly consider some points in the shape of the player and ball as the features, and choose the right algorithm (e.g., Point Distribution Model, Kalman filter, Particle filter) to associate those points through consecutive frames (see Figure~\ref{Point}). 

\begin{figure}[ht]%
    \centering
    \subfloat[Players features from frames 0, 28, 48]{{\includegraphics[height=3cm]{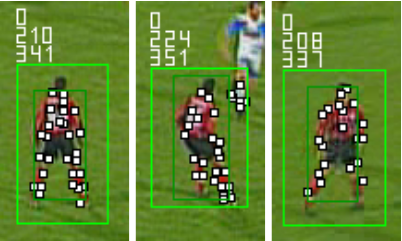} }}%
    \qquad
    \subfloat[PDM corresponding to the normalized players' shape]{{\includegraphics[height=3cm]{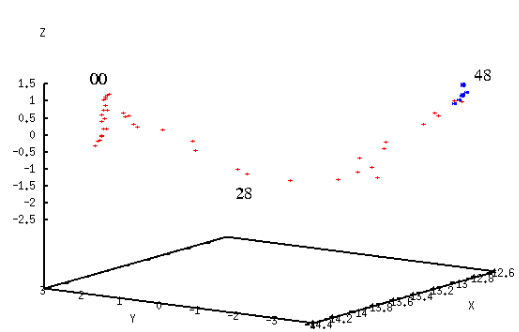} }}%
    \captionsetup{justification=centering}
    \caption{Describing shape by PDM from \citet{Mathes2006}}%
    \label{pdm}%
\end{figure}

\textbf{Point Distribution Model:}
In these methods, the idea is to describe the statistical models of the shape of players and ball, called Point Distribution Model \uppercase{(pdm)}. This method is used by several studies such as \citet{Mathes2006, Hayet2005, Li2012}. The shape is interpreted as the geometric information of the player, which is the residue once location and scaling are removed. As the first step, they extract the vector of features using 2 methods: Harris detector, or Scale Invariant Feature Transform (SIFT). Harris detector is the corner detection operator to extract corners and infer features of an image. Example results of Harris detector are shown as some points in Figure~\ref{Point}. SIFT is a feature detector algorithm to describe local features in images. These extracted features are detectable even under modifications in scale, noise, and illumination.
Then, by learning the spatial relationships between these points, they construct the PDM to concatenate all feature vectors, i.e., interest points, of players (Figure~\ref{pdm}). We provide a review and comparison of point tracking methods in Table~\ref{pointtracking}.

\begin{table}[]
\caption{Review of tracking methods with PDM} 
\centering
\begin{adjustbox}{width=1.\textwidth,center=\textwidth}

\small
\label{pointtracking}
\begin{tabular}{|>{\centering\arraybackslash}m{1.5cm}|>{\centering\arraybackslash}m{2cm}|>{\centering\arraybackslash}m{1.5cm}|>{\centering\arraybackslash}m{3cm}|>{\centering\arraybackslash}m{4.5cm}|}
  \hline
\textbf{ Reference} & \textbf{Tracking Method} &  \textbf{Point Extraction Method} & \textbf{Input video stream} & \textbf{Evaluation} 
\\ 
  \hline
  \citet{Li2012} &Features tracking in consecutive frames & SIFT features &  Football video with multiple stationary cameras & Average reliability of tracking, i.e., the number of correctly tracked players divided by the number of players in each frame is 99.7\% ; Occlusion can be handled by comparing different viewpoints of cameras\\
  \hline
  \citet{Hayet2005}  & Matching points of the PDM & Harris detector &  Football video broadcast &  Copes with the problem of rotating \& zooming cameras by continuous image-to-model homography estimation; Occlusion can be handled by interpolation in the PDM\\
  \hline
  \citet{Mathes2006} &  Points matching by maximum-gain using Hungarian algorithm & Harris detector &  Football video broadcast & Can only track the non-rigid but textured objects in crowded scenes; Occlusion can be handled by tracking sparse sets of local features\\
  \hline
\end{tabular}
\end{adjustbox}
\end{table} 



\textbf{Particle filter:} All particle filter tracking systems aim to estimate the state of a system ($x_t$), given a set of noisy observations ($z_{1:t}$). Thus the goal is to estimate $P(x_t | z_{1:t})$. If we consider this problem as a Markov process, the solution can be found if the system is assumed to be linear and each conditional probability distribution is being modeled as a Gaussian. However, these assumptions cannot be made, as they decrease the accuracy of prediction. Particle filtering can help to eliminate the necessity of extra assumptions. This method approximates the probability distribution with a weighted set of N samples:

\begin{equation} \label{eq5}
P(x) \sim \sum_{i=1}^N \omega^i (x-x_i),  
\end{equation}
where $\omega^i$ is the weight of the sample $x_i$. Now the questions are how to assign the weights, and how to sample the particles. Several studies suggested different methods for these questions.

\begin{figure}[ht]%
    \centering
    \subfloat[Set of 500 particles for $P(x_t,y_t|y_t)$ of a player]{{\includegraphics[width=4cm]{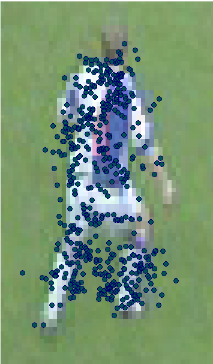} }}%
    \qquad
    \subfloat[Posterior probability distribution function given the current state of a particle $P(y_t|x_t = x)$. Darker points represent higher probability]{{\includegraphics[width=5cm]{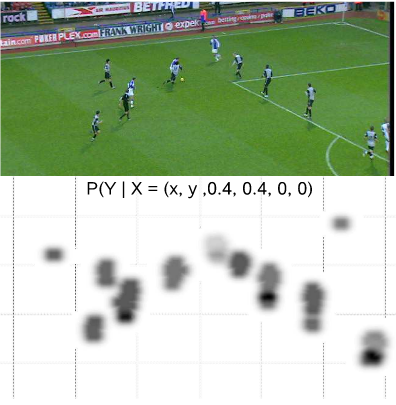} }}%
    \captionsetup{justification=centering}
    \caption{Particle filtering from \citet{Anthony2006}}%
    \label{particle}%
\end{figure}

In the methods by \citet{Kataoka2011, Manafifardd2017}, particles are players' positions. Linear uniform motion is used to model the movement of particles, and the Bhattacharyya coefficient is applied for assigning weights, i.e., likelihood to each particle. In statistics, Bhattacharyya coefficient (BC) is a measure for the amount of overlap between two statistical samples $(p,q)$ over the same domain $x$, and is calculated as $BC(p,q)= \sum_x \sqrt{p(x)q(x)}$.
In the works by \citet{Panagiotis2016, Yang2017}, each particle is estimated by the updated location of the player, knowing the last location plus a noise: $ x_k = x_{k-1} + v_k $, which noise $v_k$ is assumed to be i.i.d. following a zero-mean Gaussian distribution. Moreover, in \citet{Yang2017}, particles are created based on color and edge features of players, and the weight of each particle is computed by contrast to the similarity between the particles and targets. \citet{Anthony2006} introduced Sample Importance Resampling to show that the shape of a player can be represented by a set of particles, e.g., edge, center of mass, and color pixels. Also, those points can represent a probabilistic distribution of the state of the player (Figure~\ref{particle}). 
Another method is proposed by \citet{Pedro2015}, in which players are detected by adaptive background subtraction method based on a mixture of Gaussians, and each detected player is automatically tracked by a separate particle filter and weighted average of particles. We show the above-mentioned methods for particle filtering in Table~\ref{particlefiltering}. 

\begin{table}[ht!]
\caption{Review of particle filtering methods} 
\centering
\begin{adjustbox}{width=1.\textwidth,center=\textwidth}
\small
\label{particlefiltering}
\begin{tabular}{|>{\centering\arraybackslash}m{2cm}|>{\centering\arraybackslash}m{2cm}|>{\centering\arraybackslash}m{2cm}|>{\centering\arraybackslash}m{3cm}|>{\centering\arraybackslash}m{4cm}|}
  \hline
\textbf{ Reference} & \textbf{Particles type} &  \textbf{Weight assignment method} & \textbf{Input video stream} & \textbf{Evaluation}\\ 
  \hline

  \citet{Kataoka2011} &  Players' position \& center of gravity &  Bhattacharyya coefficient &  Football video with single swing motion camera  & Tracking rate for players: 83\% \& ball: 98\%; Occlusion handling by combining particle filter and real AdaBoost\\
  \hline
  \citet{Panagiotis2016} & Players' position & Weighted average of particles & Football video with single stationary camera & Not real-time; Occlusion cannot be handled\\
  \hline
  \citet{Yang2017} & Color \& edge features & Bhattacharyya coefficient &  Football video broadcast &  Occlusion handling by comparing color \& edge features \\
  \hline
  \citet{Anthony2006} &  Edge points, center of mass, color pixels & Sample Importance Resampling & Football video from single moving camera  & Overcomes the problem of non-linear and non-Gaussian nature of the noise model\\
  \hline

  \citet{Manafifardd2017} & Ellipse surrounded by the player bounding box & Bhattacharya coefficient &   Football video broadcast & 92\% of accuracy; occlusion can be handled by combination of particle swarm optimization \& multiple hypothesis tracking\\
  \hline

\end{tabular}
\end{adjustbox}
\end{table}

\begin{figure}[]
        \center{\includegraphics[scale=0.5]{ 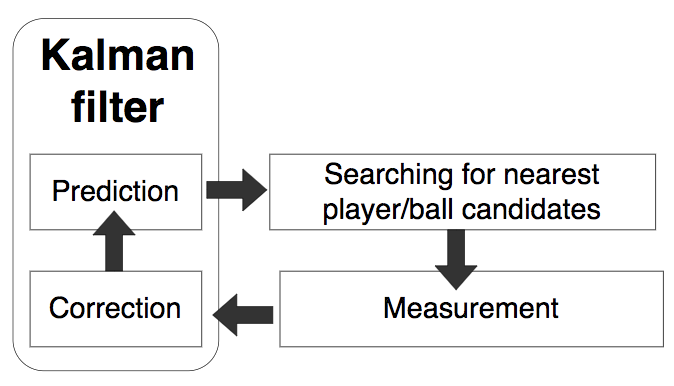}}
        \captionsetup{justification=centering}
        \caption{\label{kf}  Typical Kalman filter process}
      \end{figure}

\textbf{Kalman filter:} (KF) method is mostly used in systems with the state-space format. In the state-space models, we have a set of states evolving over time. However, the observations of these states are noisy and we are sometimes unable to directly observe the states. Thus, state-space models help to infer information of the states, given the observations, as new information arrives.  In player and ball tracking, the observations of two inputs, i.e., time and noisy position measurements, continuously update the tracker. The role of KF is to estimate the $x_t$, given the initial estimate of $x_0$, and time-series of measurements (observations), $z_1, z_2, ..., z_t$. The KF process defines the evolution of state from time $t-1$ to $t$ as:

\begin{equation} \label{eq6}
x_t = Fx_{t-1} + Bu_{t-1} + \omega_{t-1},
\end{equation}
where $F$ is the transition matrix for state vector $x_{t-1}$, $B$ is the control-input matrix for control vector $u_{t-1}$, and $\omega_{t-1}$ is the noise following a zero-mean Gaussian distribution. A typical KF process is shown in Figure~\ref{kf}. As we can see, the Kalman filter and particle filter are both recursively updating an estimate of the state, given a set of noisy observations. Kalman filter performs this task by linear projections \eqref{eq6}, while the Particle filter does so by estimating the probability distribution \eqref{eq5}. 

The following studies use Kalman filter for player and ball tracking: \citet{Aziz2018, Jong2009, Liu2011}. We summarize the KF methods in Table~\ref{KF}.

\begin{table}[ht!]
\caption{Summary of player and ball tracking methods with Kalman filter} 
\centering
\begin{adjustbox}{width=1.\textwidth,center=\textwidth}
\small
\label{KF}
\begin{tabular}{|>{\centering\arraybackslash}m{1.5cm}|>{\centering\arraybackslash}m{1.5cm}|>{\centering\arraybackslash}m{2cm}|>{\centering\arraybackslash}m{3cm}|>{\centering\arraybackslash}m{4cm}|}
  \hline
\textbf{ Reference} & \textbf{KF type} & \textbf{KF inputs} & \textbf{Input video stream} & \textbf{Evaluation}\\ 
  \hline

  \citet{Aziz2018} &  KF with motion information & Players motion information (moving or static) &  Volleyball video broadcast & Non-linear \& non-Gaussian noise are ignored, which decreases the accuracy of tracking \\
  \hline
  \citet{Jong2009} &  Dynamic KF & Position \& velocity of state vector &  Football video broadcast  & Copes with the problem of player-ball occlusion in KF\\
  \hline
  \citet{Liu2011} & Kinematic model of KF & Position state from Mean-Shift algorithm &  Basketball video broadcast & KF is used to confirm the target location to empower Mean-shift algorithm for ball tracking  \\
  \hline

\end{tabular}
\end{adjustbox}
\end{table}

\subsubsection{Contour tracking}
\begin{figure}[!htb]
        \center{\includegraphics[height=3cm]{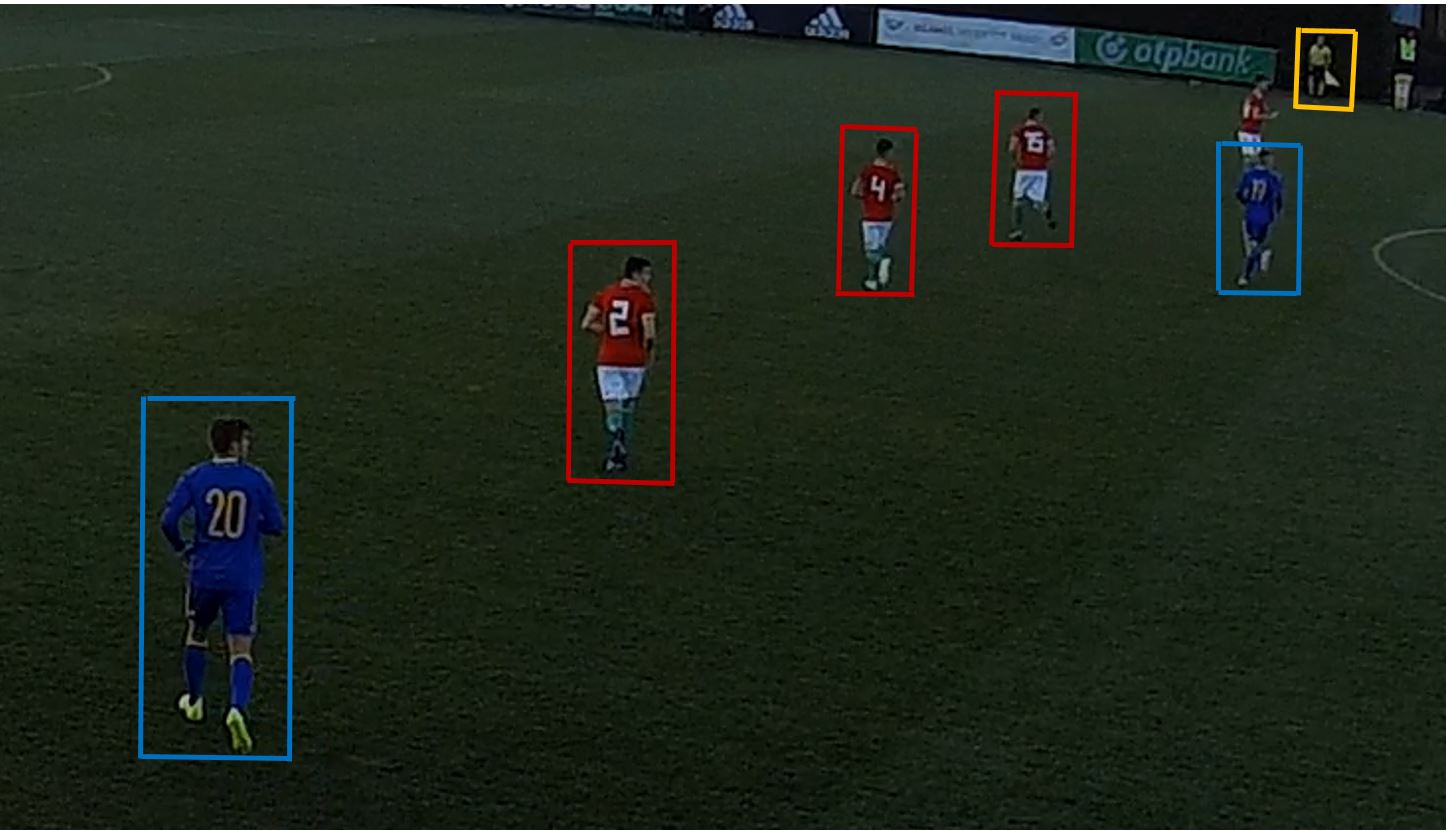}}
        \captionsetup{justification=centering}
        \caption{\label{contour} Contour tracking}
\end{figure}

\begin{figure}[!htb]
        \center{\includegraphics[height=4cm]{ 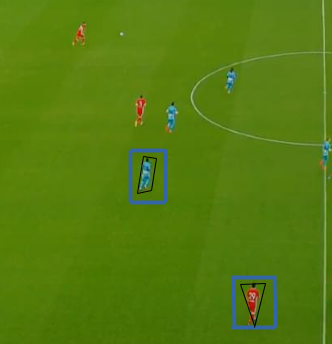}}
        \captionsetup{justification=centering}
        \caption{\label{snake1} Active contour model for fitting curves to the players' edges and lines}
      \end{figure}

Contour tracking for dynamic sports videos provides basic data, such as orientation and position of the players, and is used when we have deforming objects, i.e., players and ball, over consecutive frames. Figure~\ref{contour} shows some examples of such contours. 
Many methods have been proposed to track these contours. In an easy approach,  the centroid of these contours plus the bounding box of players will be obtained, and the player can be traced \citet{Bikramjot2013, Michael2007}.
Researchers in this area tried to propose several methods for assigning a suitable contour to the players and the ball. \citet{Pratik2018} find player's contours as curves, joining all the continuous points (along the boundary), having the same color or intensity. So they could track these contours and decide whether the player is in an offside position or not. 
Another method by \citet{Sebastien2000, Sebastien2002, Maochun2017} suggests snake or active contour tracking, which does not include any position prediction. In such methods, the algorithm fits open or close splines (i.e., a special function defined piecewise by polynomial) to lines or edges of the players. An active contour can be represented as a curve: $[x_t , y_t], t \in [0,1]$ segmenting players from the rest of the image, which can be closed or not. Then this curve should be iteratively deformed and converged to target contour (Figure~\ref{snake1}) to minimize an energy function and to fit the curve to the lines or edges of the players. The energy function is presented as physical properties of the contours, i.e., the shape of the contour, plus the gradient and intensity of the pixels in the contour. A review of contour representation of the above-mentioned tracking methods is in Table~\ref{contourtracking}.
 
 \begin{table}[ht!]
\caption{Summary of contour tracking methods} 
\centering
\begin{adjustbox}{width=1.\textwidth,center=\textwidth}
\small
\label{contourtracking}
\begin{tabular}{|>{\centering\arraybackslash}m{1.5cm}|>{\centering\arraybackslash}m{2cm}|>{\centering\arraybackslash}m{2cm}|>{\centering\arraybackslash}m{4cm}|>{\centering\arraybackslash}m{4cm}|}
  \hline
\textbf{Reference} & \textbf{Contour representation} & \textbf{Tracking method} & \textbf{Input video stram} & \textbf{Evaluation}\\ 

  \hline
  \citet{Pratik2018} &	 A curve that joins all continuous pixels &	Contour filtering for players \& ball with Gaussian blurring & Football video with multiple stationary cameras  & Fast tracking; Occlusion can be handled by placing cameras on both sides of the field\\
  \hline
  \citet{Sebastien2000,Sebastien2002} &	 Snake initialization & Snake deformation &  Football video with single moving camera & Robustly solves occlusion\\
  \hline
  \citet{Bikramjot2013} &	Edge pixels form contour boundaries & Contour centroid tracking & Football video with 3 stationary and 1 moving cameras & Handles occlusion by comparing contour area of player \& mean of that for all players\\
  \hline
  \citet{Michael2007} &   K-means clustering of pixels on marked regions & Multiple Hypothesis Tracker & Football video broadcast & Tracks players up to 20 minutes without getting lost; Detection rate is over 90\%\\
  \hline
  \citet{Maochun2017} &	Motion curve of shooting arm & Iterative convergence of dynamic contour with Lagrange equation & Basketball video broadcast & Occlusion can be handled by minimizing the potential energy of the system image\\
  \hline

  \end{tabular}
\end{adjustbox}
\end{table}

\subsubsection{Silhouette tracking}
\begin{figure}[!htb]
        \center{\includegraphics[width=7cm]{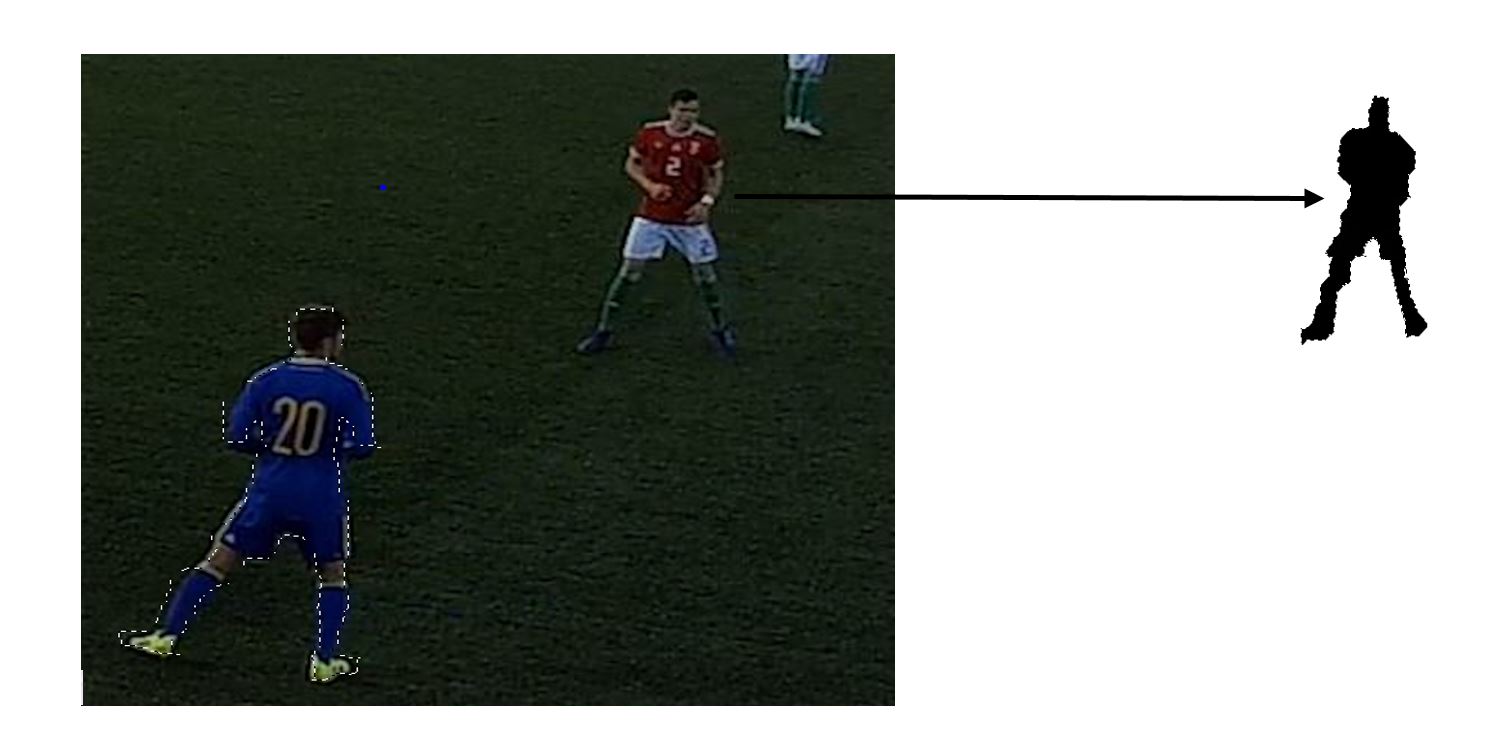}}
        \captionsetup{justification=centering}
        \caption{\label{silhouette} Silhouette tracking}
      \end{figure}

When the information provided by contour and simple geometric shapes are not enough for the tracking algorithm, extracting the silhouette of the players and of the ball can provide extra information on the appearance of the object in consecutive frames. Unlike contours, the silhouette of a player is not a curved shape. Thus, it does not require deformation and convergence to the target shape of players and the ball. Instead, this method proposes some aspect ratios to describe the invariant shape. An example of this shape extraction for a specific player is illustrated in Figure~\ref{silhouette}. In such cases, shape analysis can help the tracking process as follows.
 
\begin{figure}[!htb]
        \center{\includegraphics[height=2cm]{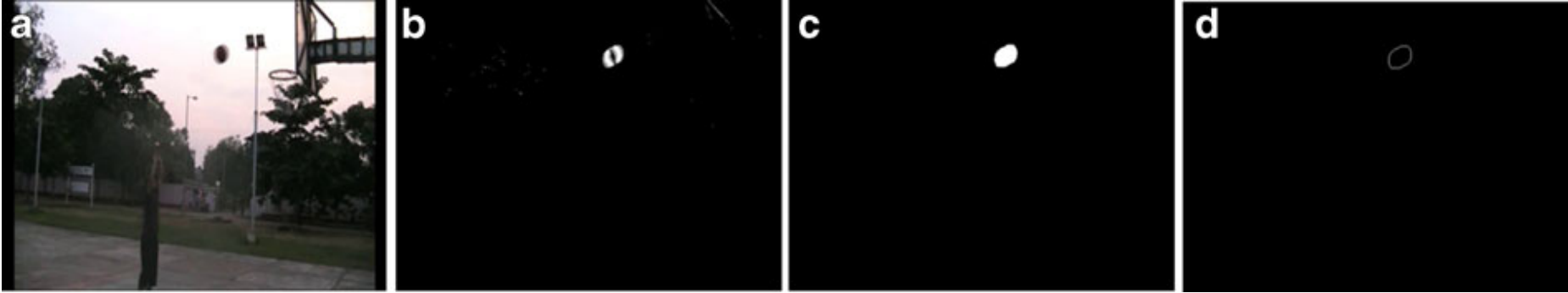}}
        \captionsetup{justification=centering}
        \caption{\label{shape} Shape of the moving ball from \citet{Bodhisattwa2012}}
      \end{figure}

\textbf{Shape matching:}
In the literature, the shape of an object is defined by its local features not determined or altered by additive contextual effects, e.g., location, scale and rotation. This method is mostly used for ball tracking. The problem in this area is that the shape of the ball varies significantly in each frame, and does not look like a circle at all (Figure~\ref{shape}). Different studies suggest some aspect ratios, i.e, shape descriptors, to get the near-circular ball images. \citet{Bodhisattwa2013} suggest using the degree of compaction $C_d$ which is the ratio of the square of the perimeter of the given shape to the area of the given shape: $C_d = P^2 / 4\pi A  $. Therefore, if $C_d > 50\%$, the shape can be filtered as a ball. Another shape descriptor is eccentricity, proposed by \citet{Wayne2006}, and it is defined as the ratio of the longest diameter to the shortest diameter of a shape. The form factor indicates how circular an object is, and if the result is between [0.2,0.65] they will consider it as a ball. Besides these shape descriptors, \citet{Huang2007} proposed using skeletons to separate a shape’s topological properties from its geometries. To extract the skeleton for every foreground blob, they use the Euclidean distance transform. Table~\ref{shapeanalysis} shows a review of shape analysis in player and ball tracking methods.

\begin{table}[ht!]
\caption{Summary of shape analysis in player and ball tracking methods} 
\centering
\begin{adjustbox}{width=1.\textwidth,center=\textwidth}
\small
\label{shapeanalysis}
\begin{tabular}{|>{\centering\arraybackslash}m{1.5cm}|>{\centering\arraybackslash}m{3cm}|>{\centering\arraybackslash}m{3cm}|>{\centering\arraybackslash}m{4cm}|}
  \hline
\textbf{ Reference} &  \textbf{Tracking method} & \textbf{Input video stream} & \textbf{Evaluation}\\ 
  \hline

  \citet{Bodhisattwa2013} & Shape, size and compaction filtering &  Basketball video broadcast & 93\% of accuracy of ball detection and tracking; Occlusion can be handled by trajectory interpolation with regression analysis\\
  \hline
  \citet{Wayne2006} & Moore-Neighbour tracing algorithm & Football video with single stationary camera & Shape analysis in this method is failing in case of the shadow of players or the ball\\
  \hline
  \citet{Huang2007}	&   Euclidean distance transform & Football video broadcast &	Occlusion cannot be handled \\
  \hline

\end{tabular}
\end{adjustbox}
\end{table}

\subsubsection{Graph-based tracking}
\begin{figure}[!htb]
        \center{\includegraphics[scale=0.5]{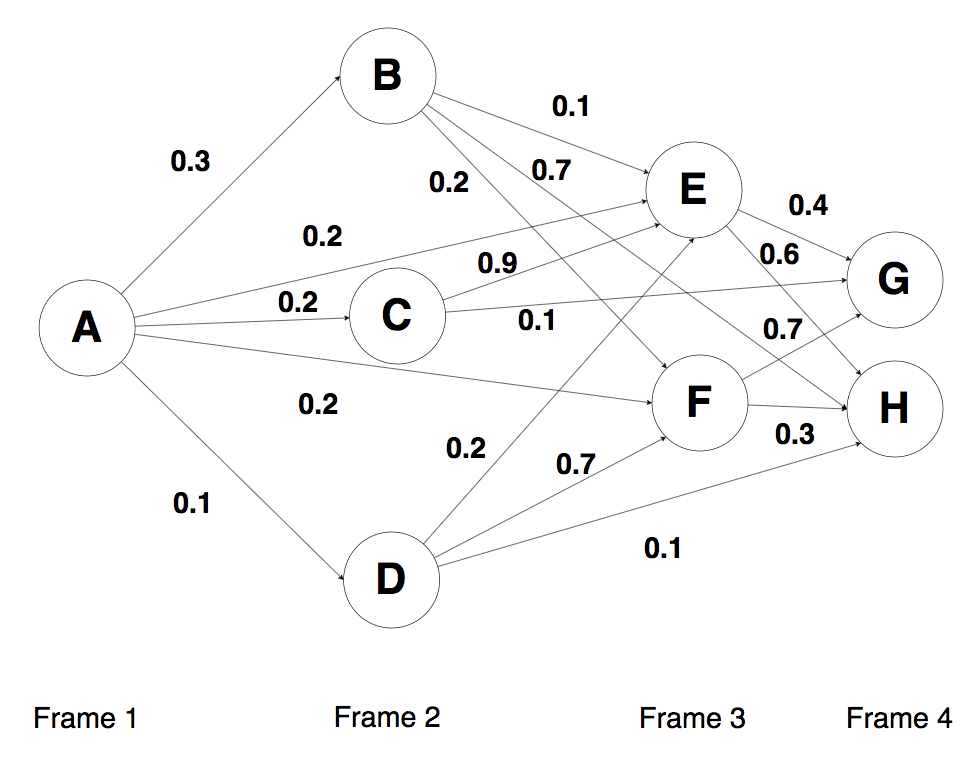}}
        \captionsetup{justification=centering}
        \caption{\label{graph} An example of weighted graph for player tracking in 4 consecutive frames}
      \end{figure}

Some works explore graph-based multiple-hypothesis to perform player tracking. In these cases, a graph is constructed that shows all the possible trajectories of players, and it models their positions along with their transition between frames. 
The correct trajectory is found with the help of, e.g., similarity measure, linear programming, multi-commodity network flow, or the problem is modeled as a minimum edge cover problem. An example of graph tracking in consecutive frames is shown in Figure~\ref{graph}.
The method shown by \citet{Figueroa2004} builds the graph in such a way that nodes represent blobs and edges represent the distance between these blobs. Then tracking of each player is performed by searching the shortest path in the graph. However, occlusion is difficult to be handled with this method. 
Authors of \citet{Pallavi2008} used dynamic programming to find the optimal trajectory of each player in the graph.
The proposed method by \citet{Junliang20011} builds an undirected graph to model the occlusion relationships between different players.
In \citet{Chen2017}, the method constructs a layered graph for detected players, which includes all probable trajectories. Each layer corresponds to a frame and each node represents a player. Two nodes of adjacent layers are linked by an edge if their distance is less than a pre-defined threshold.
Finally, the authors used the Viterbi algorithm in dynamic programming to extract the shortest path of the graph. 
Ball tracking with graphs was proposed in \citet{Andrii2015}, where they build a ball graph to formulate the Mixed Integer Programming model, and each node is associated with a state, i.e., location of the ball at a time instance. Table~\ref{graphbased} shows a review of node and edge representation, along with tracking methods defined on the graph.

\begin{table}[ht!]
\caption{Summary of graph-based player and ball tracking methods} 
\centering
\begin{adjustbox}{width=1.\textwidth,center=\textwidth}
\small
\label{graphbased}
\begin{tabular}{|>{\centering\arraybackslash}m{1.5cm}|>{\centering\arraybackslash}m{1.5cm}|>{\centering\arraybackslash}m{2cm}|>{\centering\arraybackslash}m{2cm}|>{\centering\arraybackslash}m{3cm}|>{\centering\arraybackslash}m{4cm}|}
  \hline
\textbf{Reference} &  \textbf{Node representation} & \textbf{Edge representation} & \textbf{Tracking method} & \textbf{Input video stream} & \textbf{Evaluation}\\ 
  \hline
  \citet{Figueroa2004}  & Blobs & Distance between blobs & Minimal path of graph & Football video with multiple stationary cameras & The algorithm is tested for 3 players of defender, mid-fielder, and forwarder, and shows 88\% of solved occlusions\\
  \hline
  \citet{Pallavi2008} &  Probable player candidates & Candidates link between frames & Dynamic programming with acyclic graph &  Football video broadcast & 93\% of accuracy in tracking \& 80\% of solved occlusions\\
 \hline
  \citet{Junliang20011} & Player & Relationship ratio between 2 players & Dual-mode two-way Bayesian inference approach &  Football and basketball video broadcast  & Uses undirected graph to model the occlusion relationships \& reports 119 Mostly Tracked trajectories \& 12 ID switches \\
  \hline
  \citet{Chen2017} &  Players position & Degree of closeness between players &  Viterbi algorithm to find shortest path &  Basketball video broadcast &	88\% of precision in player tracking; Occlusion is handled by layered graph connections\\
  \hline
  \citet{Andrii2015} &  Ball's location & State-Time instant connection & Mixed Integer Programming &	Football, volleyball, and basketball video with multiple cameras & 97\% of accuracy in player tracking \& 74\% in ball tracking \\
  \hline

\end{tabular}
\end{adjustbox}
\end{table}

\subsubsection{Data association methods}

Simulation-based approaches, including Monte Carlo methods and joint probabilistic data association, are usually used for solving multitarget tracking problems, as these methods perform well for nonlinear and non-Gaussian data models. 


\textbf{Markov Chain Monte Carlo data association \uppercase{(mcmc)}:}
\citet{Septier2011} compared several \uppercase{mcmc} methods, such as 1) sequential importance resampling algorithm, 2) resample-move, 3) \uppercase{mcmc}-based particle method. The difference between these methods stems from the sampling strategy from posterior by using previous samples. Simulations show that the \uppercase{mcmc}-based Particle approach exhibits better tracking performance and thus clearly represents interesting alternatives to Sequential Monte Carlo methods. The authors of \citet{Liu2009} designed a Metropolis-Hastings sampler for \uppercase{mcmc}, which increased the efficiency of the method.

\textbf{Joint probabilistic data association \uppercase{(jpda)}:}
The JPDA method can be used when the mapping from tracks to observations is not clear, and we do not know which observations are valid and which are just noise. In these cases, \uppercase{jpda} implements a probabilistic assignment. \citet{Robert2009} used \uppercase{jpda} 
to assign the probability of association between each observation and each track.

\subsection{Deep learning-based tracking}
\label{sec:dee}

Despite the effectiveness of traditional methods, they fail in many real-world scenarios, e.g., occlusion, and processing videos from several viewpoints. On the other hand, deep learning models benefit from the learning ability of neural networks on large and complex datasets, and they eliminate the necessities of features extraction by the human/expert. Therefore, deep learning-based trackers are recently getting much attention in computer vision. These trackers are categorized into online and offline methods: online trackers are trained from scratch during the test and are not taking advantage of already annotated videos for improving performance, while offline trackers train on offline data.

Several recent studies have attempted to assess the performance of deep learning methods in sports analytics. The core idea of all methods is to use CNN. However, each study proposes a different structure of the network and training method for increasing the performance. In this section, we summarize the state-of-the-art networks and their application in sports analytics. Table~\ref{table:deep} is a brief review of these methods.

\begin{figure}[!htb]
        \center{\includegraphics[width=10cm]{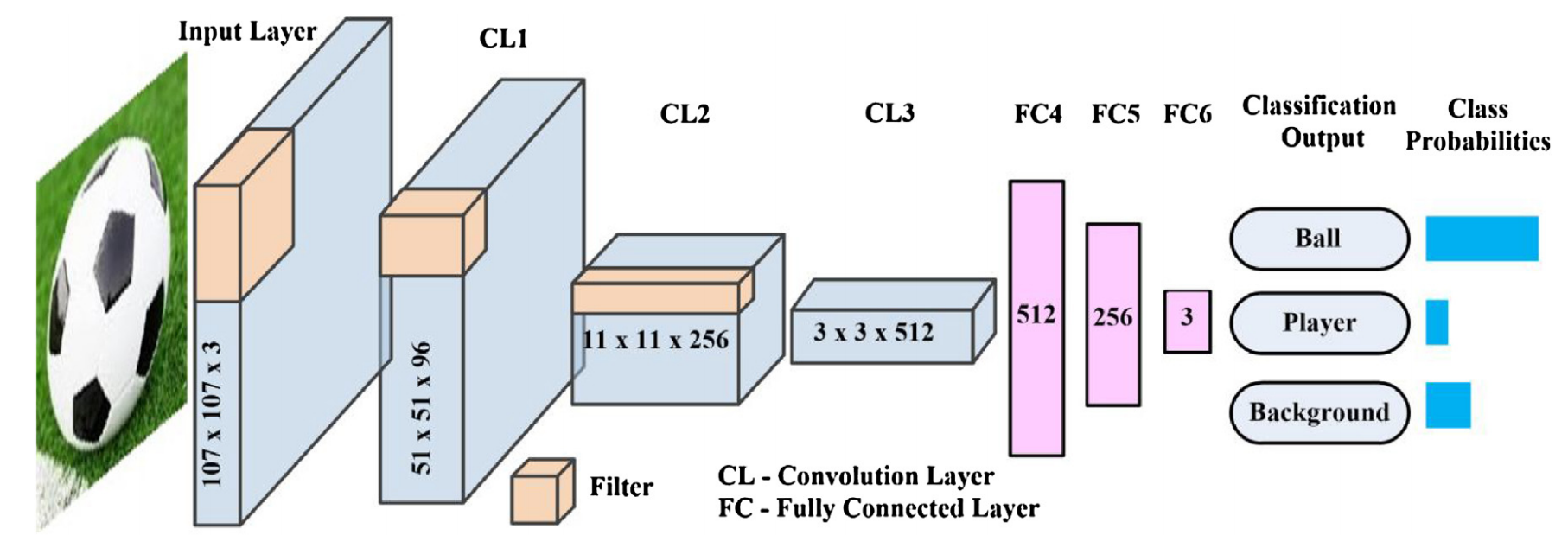}}
        \captionsetup{justification=centering}
        \caption{\label{vggm} VGG-M CNN architecture from \citet{Kamble2019}}
      \end{figure}

\textbf{Visual Geometry Group (VGG): }VGG-M is a CNN architecture, designed by the Visual Geometry Group (VGG) at the University of Oxford. This network is used by several studies such as \citet{Kamble2019, Adria2019}. VGG-M is a small type of CNN, and its pre-trained weights are publicly available. This network gets the image as input, and classifies the detected object as player, ball, or background, along with the probability of the classes. The architecture of VGG-M CNN is illustrated in Figure~\ref{vggm}.

After the classification of the players and ball, the metric called Intersection Over Union (IOU) is used to track them. IOU is the ratio of intersection of the ground truth bounding box from the previous frame ($BB_A$), and predicted bounding box in the current frame ($BB_B$), and it is calculated as in \eqref{eq7}:

\begin{equation} \label{eq7}
IOU= \frac{|BB_A \cap BB_B|}{|BB_A \cup BB_B|},
\end{equation}
where $\cap$ and $\cup$ are intersection and union in terms of the number of pixels. Thus, if the intersection is non-zero between consecutive frames, the player or ball can be traced.

\begin{figure}[!htb]
        \center{\includegraphics[width=8cm]{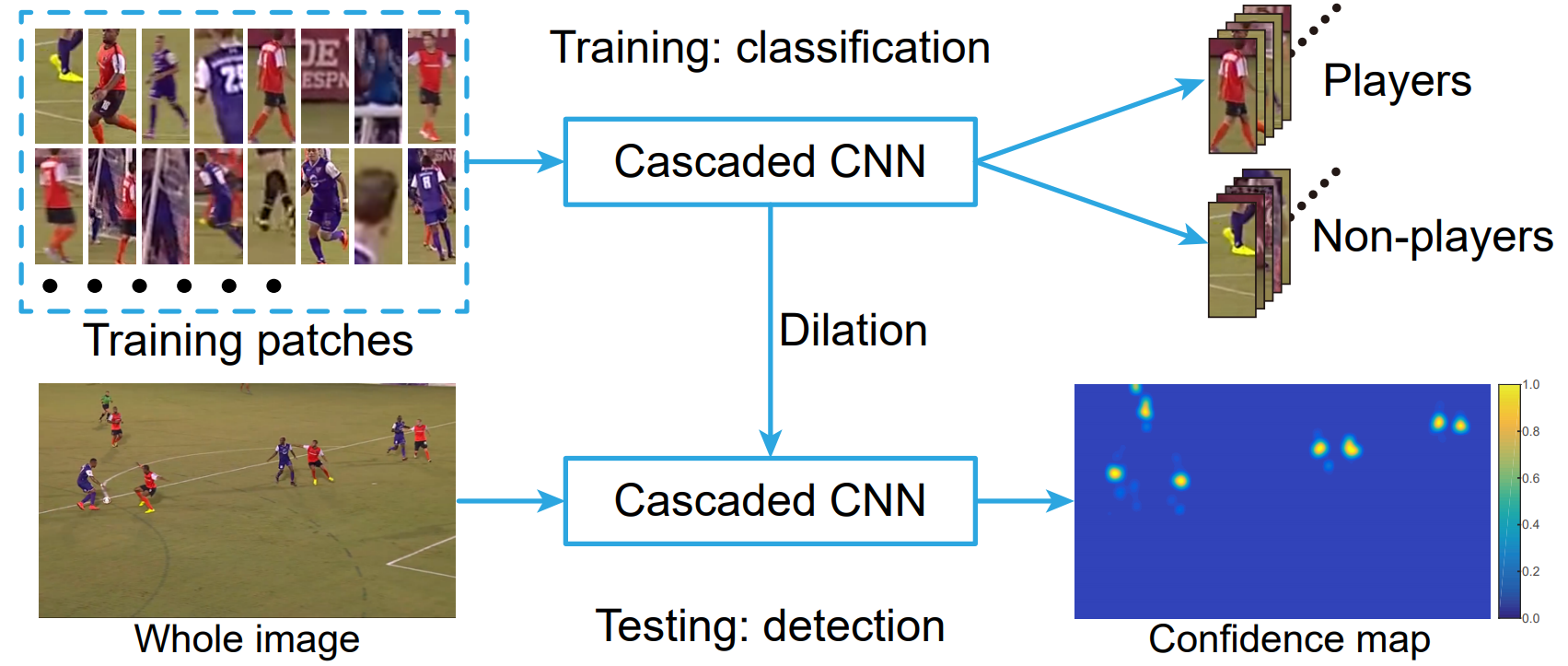}}
        \captionsetup{justification=centering}
        \caption{\label{ccnn} Classification process with Cascade-CNN from \citet{KeyuLu2017}}
      \end{figure}

\textbf{Cascade-CNN: } is a novel deep learning architecture consisting of multiple CNNs. This network is trained on labeled image patches and classifies the detected objects into the two classes of player and non-player. Football and basketball player tracking using this method is suggested by \citet{KeyuLu2017}. The illustrated pipeline in Figure~\ref{ccnn} shows the classification process and a dilation strategy for accurate player tracking with the help of IOU metric. 

\begin{figure}[ht]%
    \centering
    \subfloat[\centering Player and ball classification with YOLO]{{\includegraphics[ width=10cm]{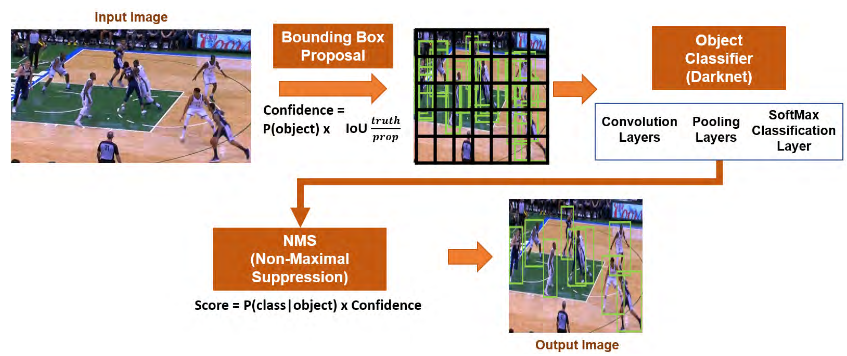} }}%
    \qquad
    \subfloat[\centering IOU evaluation for tracking]{{\includegraphics[width=2cm]{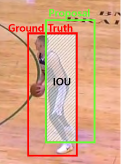} }}%
    \captionsetup{justification=centering}
    \caption{Player and ball tracking with YOLO  from \citet{Young2019}}%
    \label{yolo}%
\end{figure}

\textbf{YOLO: } This network is used by \citet{Matija2019} for handball player and ball tracking, and \citet{Young2019} for basketball player movement recognition. YOLO applies a single neural network to the full image. Then the network divides the image into cells and predicts bounding boxes and probabilities for each cell. The weights of the bounding boxes are the predicted probabilities. Then IOU metric can help for tracking purposes and solving the occlusion problem of the players and the ball (Figure ~\ref{yolo}).    

\begin{figure}[!htb]
        \center{\includegraphics[width=5cm]{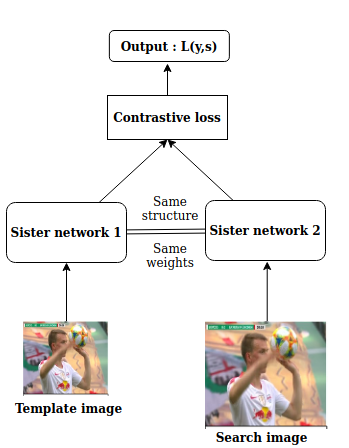}}
        \captionsetup{justification=centering}
        \caption{\label{siam} SiamCNN network structure for player tracking}
      \end{figure}

\textbf{SiamCNN: } In this network, there are sister network 1 and sister network 2 with the same network structure, parameters, and weights. The structure looks like VGG-M except for the adjustment of the sizes of each layer. The inputs of SiamCNN are 3-color channels (R,G,B) from frames, and the output is the Euclidean distance between the characteristics/features of the inputs. \citet{Long2019} used this network to extract players' characteristics through trajectories. Then they compare the similarities between search areas and a target template, so players can be tracked. The structure of this network is given in Figure~\ref{siam}.

\begin{table}[]
\caption{Summary of deep learning methods and application in team sports (AUC: Area Under Curve; mAP: mean Average Precision; MAPE: Mean Absolute Percentage Error; MOTA: Multi Object Tracking Accuracy)}
\centering
\begin{adjustbox}{width=1.\textwidth,center=\textwidth}
\small
\label{table:deep}
\begin{tabular}{|>{\centering\arraybackslash}m{1.5cm}|>{\centering\arraybackslash}m{2cm}|>{\centering\arraybackslash}m{4cm}|>{\centering\arraybackslash}m{3.5cm}|>{\centering\arraybackslash}m{4cm}|}
  \hline
  \textbf{Reference} & \textbf{Network structure } & \textbf{Input video stream} & \textbf{Required computational resource(s)} & \textbf{Performance}\\ 
  \hline

  \citet{KeyuLu2017} & Cascade CNN & Football and basketball video broadcast & Intel i7-6700HQ; NVIDIA GTX1060 & AUC of player detection is 0.97\\\hline
  
  \citet{Kamble2019} &  VGG-M &   Football video with multiple stationary cameras & MATLAB 2018a; Intel i7; NVIDIA GTX1050Ti & 87\% of accuracy in player, ball, and event detection \\\hline
  \citet{Long2019} & Full-convolution Siamese NN & Football video broadcast & Matlab2014a; Intel i7; NVIDIA GTX 960 M & Mean value of target tracking effect of SiamCNN is 60\% \\\hline

  \citet{Young2019} & YOLO & Basketball video with single moving camera & Intel i7; NVIDIA GeForce GTX 1080Ti & 74\% of precision in recognizing Jersey numbers; MAPE is at most 34\% \\\hline
  \citet{Adria2019} & VGG-19 & Basketball video with single moving camera & - & Detection precision: 98\% ; MOTA of tracking: 68\%\\\hline
  
  \citet{Matija2019} & YOLO & Handball video with multiple stationary cameras & 12 core E5-2680v3 CPU; GeForce GTX TITAN & mAP in players \& ball detection: 37\%\\\hline

\end{tabular}
\end{adjustbox}
\end{table}

\section{Evaluation and model selection}
\label{sec:eval}

\begin{figure}[ht]%
    \centering
    \subfloat[\centering Detection task]{{\includegraphics[height=5cm]{ 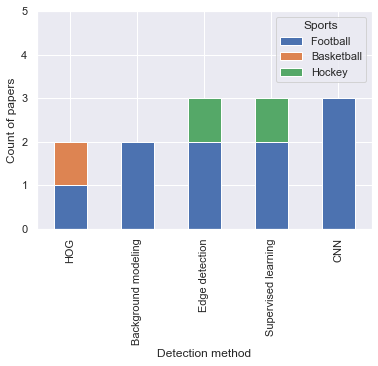} }}%
    \qquad
    \subfloat[\centering Tracking task]{{\includegraphics[height=5cm]{ 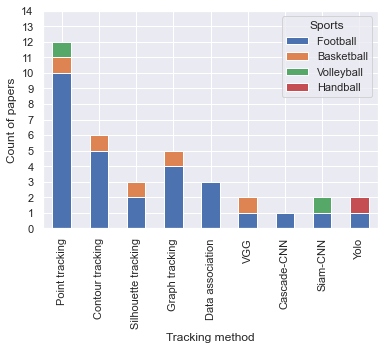} }}%
    \captionsetup{justification=centering}
    \caption{Number of the published papers for each method categorized by their application in team sports}%
    \label{methods}%
\end{figure}

If a clean set of tracking information is not provided to a sports analyzer who is developing a quantitative model, his/her core task is to choose the most suitable method for tracking players and the ball, and construct the required dataset for further analysis. In the detection and tracking domains, model selection, i.e., DL or traditional, heavily depends on the task at hand. The selection will be difficult by merely reviewing the performance metrics of the methods, as the tracking performance relies on the specific task at hand and the quality of the videos. However, there are some concrete criteria in this domain, which can help the analyst to rapidly choose the desired tracking method. Figure~\ref{methods} compares the number of publications in detection and tracking domains categorized by team sports. Note that 74\% of methods are applied on football videos, whereas deep learning methods (i.e., CNN, VGG, Cascade, Siam, Yolo) are covering only 20\% of all publications. In this section, we review the benefits and drawbacks of each method, and compare them in terms of their estimated costs.

\subsection{Deep learning-based vs. traditional methods}

In general, traditional methods are domain-specific, thus the analyzer must specifically describe and select the features (e.g., edge, color, points, etc.) of the ball, football player, basketball player, background, etc. in detail. Therefore, the performance of the traditional models depends on the analyzer's expertise and how accurate the features are defined. DL methods, on the other hand, demonstrate superior flexibility and automation in detection and tracking tasks, as they can be trained offline on a huge dataset, and then automatically extract features of any object type. In this case, the necessities of manual feature extraction are eliminated, and consequently, DL requires less expertise from the analyzer. In another point of view, DL models are more like a black box on the detection tasks. On the contrary, traditional methods provide more visibility and interpretability to the analyzer on how the developed algorithm can be performed in different situations such as sports types, lighting conditions, cameras, video quality, etc. So, traditional models can give a better opportunity to improve the tracker accuracy, when the system components are visible. Also in the case of failure, system debugging are more straightforward in traditional models than DL-based ones. 

In addition to the pros and cons that are listed in this survey for each method, few criteria can help sports analysts to choose their desired method. Table~\ref{criteria} lists these criteria that can help analyzers to choose the suitable detection and tracking methods in the direction of DL-based or traditional ones.

\begin{table}[]
\caption{Model selection criteria}
\centering
\begin{adjustbox}{width=.7\textwidth,center=\textwidth}
\small
\label{criteria}
\begin{tabular}{|>{\centering\arraybackslash}m{5cm}|>{\centering\arraybackslash}m{2cm}|>{\centering\arraybackslash}m{2cm}|}
  \hline
  \textbf{Criteria} & \textbf{Deep Learning } & \textbf{Traditional} \\ 
  \hline

  Availability of huge training dataset & \checkmark &  \\\hline
  
  Accessing to high computational power & \checkmark &  \\\hline

  Lack of storage &  & \checkmark \\\hline
  
  Looking for cheaper solution & & \checkmark \\\hline
  
  Certainty and expertise in the object features  & & \checkmark \\\hline
  
  Less domain expertise & \checkmark &  \\\hline 
  
  Flexibility in terms of objects and training dataset & \checkmark &  \\\hline 
  
  Flexibility of deployment on different hardware & & \checkmark \\\hline
  
  Short training and annotation time & & \checkmark \\\hline

\end{tabular}
\end{adjustbox}
\end{table}

\subsection{Cost analysis}
\label{cost}
The cost of the method is one of the most important characteristics of model selection for researchers and analysts: they are looking for a method with maximum accuracy and reasonable cost. Here we give an insight into the cost of the state-of-the-art methods, both for infrastructure and computation, and classify them into 3 categories: high, medium, low. The classification is based on the following facts. In the computational aspect, deep learning methods which require GPUs are more expensive than traditional methods with only CPUs. From an infrastructure perspective, different methods require different sets of camera settings to record the sports video. Methods that require a set of moving or stationary camera(s) to be set up in the arena are more expensive than the methods that can trace players and the ball on broadcast video. Table~\ref{costs} shows the cost approximation of all methods along with their most significant limitations.

\begin{table}[]
\caption{Comparing cost of the methods} 
\begin{adjustbox}{width=1.\textwidth,center=\textwidth}
\label{costs}
\centering

\begin{tabular}{|>{\centering\arraybackslash}m{3cm}|>{\centering\arraybackslash}m{2cm}|>{\centering\arraybackslash}m{2cm}|>{\centering\arraybackslash}m{1.5cm}|>{\centering\arraybackslash}m{1.5cm}|>{\centering\arraybackslash}m{1.5cm}|>{\centering\arraybackslash}m{1.5cm}|>{\centering\arraybackslash}m{1cm}|>{\centering\arraybackslash}m{3cm}|}

\hline
 \textbf{Reference} & \textbf{Detection or tracking method} & \textbf{Cost approximation} &  \multicolumn{3}{c|}{\textbf{Infrastructure requirements for sport video}} & \multicolumn{2}{c|}{\textbf{Computational requirements}} & \textbf{Limitation}\\
 \hline
  & &  & \textbf{Moving camera(s)} & \textbf{Stationary camera(s)} &\textbf{ Broadcast }& \textbf{GPU} & \textbf{CPU} &  \\
 \hline
   \citet{Sebastian2015}  & deep learning & middle & & & \checkmark & \checkmark & & poor performance in perspective distortion of camera \\
\hline
  \citet{Gen2018} & deep learning & high & \checkmark & & & \checkmark & &  expensive manual force for labeling \\
  \hline
    \citet{Hengyue2019} & deep learning & high & & \checkmark & & \checkmark & & high network training time \\
  \hline
  \citet{Young2019}& deep learning & high & \checkmark & & & \checkmark & & very low accuracy \\
  \hline

  \citet{Adria2019} & deep learning & high & \checkmark & & & \checkmark & & difficult network tuning \\
  \hline

  \citet{Kamble2019} & deep learning & high & & \checkmark & & \checkmark & & manual network parameters is needed to be assigned \\
  \hline
  \citet{Matija2019} & deep learning & high & & \checkmark & & \checkmark & & -\\
  \hline

\citet{KeyuLu2017}  & deep learning & middle & & & \checkmark & \checkmark & & unrealistic uniform background color is assumed \\
  \hline
 \citet{Long2019}& deep learning & middle & & & \checkmark & \checkmark & & too sensitive on number of convolutional layers\\
  \hline
\citet{Slawomir2010} & HOG & low & & & \checkmark & & \checkmark & cannot detect occluded players \\
\hline
\citet{Evan2015}& HOG & low & & & \checkmark & & \checkmark & -\\
\hline
\citet{Ming2009} & GMM & middle & \checkmark & & & & \checkmark & performs only in absence of shadow \\
\hline
\citet{Mazzeo2008} & energy evaluation & middle &  &\checkmark & & & \checkmark & high computational time \\
\hline
 \citet{Direkoglu2018}& edge detection & middle &  &\checkmark & & & \checkmark & high computational time and only with fixed camera \\
\hline
\citet{Naushad2012,Upendra2015} & sobel gradient & low & & & \checkmark & & \checkmark & low performance in crowded places \\
\hline
\citet{Branko2015} & adaboost & middle & \checkmark & & & & \checkmark & highly training time \\
\hline
\citet{Guangyu2006} & SVM & low & & & \checkmark & & \checkmark & time consuming due to manual labeling \\
\hline
\citet{Chengjun2018} & SVM & low & & & \checkmark & & \checkmark & - \\
\hline
\citet{Li2012} & point tracking & middle & & \checkmark & & & \checkmark & - \\
\hline
\citet{Hayet2005}& point tracking& low & & & \checkmark & & \checkmark & extracted trajectories are too short \\
\hline 
\citet{Mathes2006} & point tracking& low & & & \checkmark & & \checkmark & cannot track untextured objects \\
\hline 
\citet{Kataoka2011} & particle filter & middle & \checkmark & & & & \checkmark & - \\
\hline
\citet{Panagiotis2016} & particle filter & middle &  & \checkmark & & & \checkmark & tracking fails in case of player jumping or falling\\
\hline
\citet{Yang2017} & particle filter & low & & & \checkmark & & \checkmark & inadequacy of identifying players\\
\hline 
\citet{Anthony2006} & particle filter & middle & \checkmark & & & & \checkmark & can track players only in image space, not in real-time moving camera system \\
\hline
\citet{Manafifardd2017}  & particle filter & low & & & \checkmark  & & \checkmark & players' color features are pre-selected, but they are changing in each game\\
\hline 
\citet{Pedro2015} & particle filter & low & & \checkmark &  & & \checkmark & lots of tracking id switches \\
\hline 
\citet{Aziz2018} & Kalman filter & low & & & \checkmark & & \checkmark & non-linear, non-Gaussian noises are ignored \\
\hline 
\citet{Jong2009} & Kalman filter & low & & & \checkmark & & \checkmark & this algorithm fails in the frames with crowded players \\
\hline 
\citet{Liu2011} & Kalman filter & low & & & \checkmark & & \checkmark & shot event is required for ball tracking \\
\hline 
\citet{Pratik2018} & contour tracking & middle & & \checkmark &  & & \checkmark & offside event is required for tracking \\
\hline 
\citet{Sebastien2000,Sebastien2002} & contour tracking & middle & \checkmark & & & & \checkmark & high computational time \\
\hline
\citet{Bikramjot2013} & contour tracking & middle & \checkmark & \checkmark & & & \checkmark & manual camera setting and zooming is required \\
\hline
\citet{Michael2007} & contour tracking & low & & & \checkmark & & \checkmark & - \\
\hline
\citet{Maochun2017} & contour tracking & low & & & \checkmark & & \checkmark & shooting arm is required for tracking\\
\hline
\citet{Bodhisattwa2013} & shape matching & low & & & \checkmark & & \checkmark & long shot sequences are required for ball tracking\\
\hline
\citet{Wayne2006} & shape matching & middle & & \checkmark  & & & \checkmark & this algorithm fails in shadow \\
\hline
\citet{Huang2007}& shape matching & low & & & \checkmark & & \checkmark & - \\
\hline
\citet{Figueroa2004} & graph-based & middle & & \checkmark  & & & \checkmark & -\\
\hline
\citet{Pallavi2008} & graph-based & low & & & \checkmark & & \checkmark & short tracking sequence as it focuses on solving occlusion\\
\hline
\citet{Junliang20011}& graph-based & low & & & \checkmark & & \checkmark & - \\
\hline
\citet{Chen2017}& graph-based & low & & & \checkmark & & \checkmark & -\\
\hline
\citet{Andrii2015} & graph-based & middle & \checkmark & & & & \checkmark & - \\
\hline

\end{tabular}
\end{adjustbox}
\end{table}

\section{Conclusion and future research directions}
\label{sec:ind}

According to a large number of citetd papers in this survey, computer vision researchers intensively investigate robust methods of optical tracking in sports. In this survey, we have categorized the literature according to the applied methods and video type they build on. Moreover, we elaborated on the detection phase, as a necessary preprocessing step for tracking by conventional and deep learning methods.
We believe that this survey can significantly help quantitative analysts in sports to choose the most accurate, while cost-effective tracking method suitable for their analysis. Furthermore, the combination of traditional and deep learning methods can be rarely seen in the literature. Traditional models are time-consuming and require domain expertise due to some manual feature extraction tasks, while deep learning models are quite expensive to run in terms of computing resources. As possible future work, research may aim to combine those methods to increase the performance of tracking systems, along with the robust quantitative evaluation of the games. Another avenue for future work might be to minimize the computational costs of tracking systems with the aid of sophisticated data processing methods. We hope that this survey can give an insight to sports analytics researchers to recognize the gaps of state-of-the-art methods, and come up with novel solutions of tracking and quantitative analysis.

\section*{Acknowledgment}

Project no. 128233 has been implemented with the support provided by the Ministry of Innovation and Technology of Hungary from the National Research, Development and Innovation Fund, financed under the FK\_18 funding scheme.


\bibliographystyle{unsrtnat}
\bibliography{template}  

\begin{thebibliography}{68}
\providecommand{\natexlab}[1]{#1}
\providecommand{\url}[1]{\texttt{#1}}
\expandafter\ifx\csname urlstyle\endcsname\relax
  \providecommand{\doi}[1]{doi: #1}\else
  \providecommand{\doi}{doi: \begingroup \urlstyle{rm}\Url}\fi

\bibitem[Alper et~al.(2006)Alper, Omar, and Mubarak]{yilmaz2006}
Yilmaz Alper, Javed Omar, and Shah Mubarak.
\newblock Object tracking: A survey.
\newblock \emph{\uppercase{ACM} Computing Surveys (\uppercase{CSUR})},
  38:\penalty0 45, 2006.

\bibitem[{Reddy} et~al.(2015){Reddy}, {Priya}, and {Neelima}]{RasoolReddy2015}
K.~R. {Reddy}, K.~H. {Priya}, and N.~{Neelima}.
\newblock Object detection and tracking -- a survey.
\newblock In \emph{International Conference on Computational Intelligence and
  Communication Networks (CICN)}, pages 418--421, 2015.

\bibitem[Dhenuka et~al.(2018)Dhenuka, Udesang, and Hemant]{Dhenuka2018}
M.Patel Dhenuka, K.Jaliya Udesang, and D.Vasava Hemant.
\newblock Multiple object detection and tracking: A survey.
\newblock \emph{International Journal for Research in Applied Science \&
  Engineering Technology (\uppercase{IJRASET})}, 6\penalty0 (2):\penalty0
  809--813, 2018.

\bibitem[Lee et~al.(2014)Lee, Liew, Cheah, and Wang]{Lee2014}
BY~Lee, LH~Liew, WS~Cheah, and YC~Wang.
\newblock Occlusion handling in videos object tracking: A survey.
\newblock \emph{\uppercase{IOP} Conference Series: Earth and Environmental
  Science}, 18:\penalty0 1--5, feb 2014.
\newblock \doi{10.1088/1755-1315/18/1/012097}.

\bibitem[Ciaparrone et~al.(2019)Ciaparrone, Luque-Sanchez, Tabik, Troiano,
  Tagliaferri, and Herrera]{Gioele2019}
Gioele Ciaparrone, Francisco Luque-Sanchez, Siham Tabik, Luigi Troiano, Roberto
  Tagliaferri, and Francisco Herrera.
\newblock Deep learning in video multi-object tracking: A survey.
\newblock \emph{Journal of Neurocomputing}, 4:\penalty0 1--42, 2019.

\bibitem[Manafifard et~al.(2017{\natexlab{a}})Manafifard, Ebadi, and
  Abrishami]{Manafifard2017}
M~Manafifard, H~Ebadi, and H~Moghaddam Abrishami.
\newblock A survey on player tracking in soccer videos.
\newblock \emph{Computer Vision and Image Understanding}, 159:\penalty0 19--46,
  2017{\natexlab{a}}.

\bibitem[Burke(2019)]{Bryan2019}
Bryan Burke.
\newblock {DeepQB}: Deep learning with player tracking to quantify quarterback
  decision-making \& performance.
\newblock In \emph{\uppercase{mit sloan} Sports Analytics Conference}, 2019.

\bibitem[Needham and Boyle(2001)]{Needham2001}
Chris Needham and Roger~David Boyle.
\newblock Tracking multiple sports players through occlusion, congestion and
  scale.
\newblock In \emph{Proceedings of the British Machine Vision Conference}, 2001.

\bibitem[Rodriguez-Canosa et~al.(2012)Rodriguez-Canosa, Thomas, del Cerro,
  Barrientos, and MacDonald]{Gonzalo2012}
G.~R. Rodriguez-Canosa, Stephen Thomas, Jaime del Cerro, Antonio Barrientos,
  and Bruce MacDonald.
\newblock A real-time method to detect and track moving objects
  (\uppercase{DATMO}) from unmanned aerial vehicles (\uppercase{UAVs}) ssing a
  single camera.
\newblock \emph{Remote Sensing}, 4\penalty0 (4):\penalty0 1090--1111, 2012.

\bibitem[Sabirin et~al.(2015)Sabirin, Sankoh, and Naito]{SabirinHiroshi2015}
Houari Sabirin, Hiroshi Sankoh, and Sei Naito.
\newblock Automatic soccer player tracking in single camera with robust
  occlusion handling using attribute matching.
\newblock \emph{\uppercase{IEICE} Transactions}, 98-D:\penalty0 1580--1588,
  2015.

\bibitem[Arbues et~al.(2019)Arbues, Ballester, and Haro]{Adria2019}
Adria Arbues, Coloma Ballester, and Gloria Haro.
\newblock Single-camera basketball tracker through pose and semantic feature
  fusion.
\newblock arXiv preprint, arXiv:1906.02042v2, 2019.

\bibitem[Ren et~al.(2008)Ren, Orwell, Jones, and Xu]{Jinchang2008}
Jinchang Ren, James Orwell, Graeme~A. Jones, and Ming Xu.
\newblock Real-time modeling of 3-d soccer ball trajectories from multiple
  fixed cameras.
\newblock \emph{IEEE Transactions on Circuits and Systems for Video
  Technology}, 18\penalty0 (3):\penalty0 350--362, 2008.

\bibitem[Ren et~al.(2009)Ren, Orwell, Jones, and Xu]{Jinchan2009}
Jinchang Ren, James Orwell, Graeme~A. Jones, and Ming Xu.
\newblock Tracking the soccer ball using multiple fixed cameras.
\newblock \emph{Computer Vision and Image Understanding}, 113\penalty0
  (5):\penalty0 633--642, 2009.

\bibitem[Wu(2008)]{Lan2008}
Lan Wu.
\newblock \emph{Multi-view hockey tracking with trajectory smoothing and camera
  selection}.
\newblock PhD thesis, The University Of British Columbia, 2008.

\bibitem[Yazdi and Bouwmans(2018)]{Yazdi2018}
Mehran Yazdi and Thierry Bouwmans.
\newblock New trends on moving object detection in video images captured by a
  moving camera: A survey.
\newblock \emph{Computer Science Review}, 28:\penalty0 157--177, 2018.

\bibitem[Xu et~al.(2004)Xu, Orwell, and Jones]{Ming2004}
Ming Xu, James Orwell, and Graeme Jones.
\newblock Tracking football players with multiple cameras.
\newblock In \emph{International Conference on Image Processing
  (\uppercase{ICIP})}, volume~5, pages 2909--2912, 2004.

\bibitem[Agelet~Ruiz(2010)]{Neus2010}
Neus Agelet~Ruiz.
\newblock \emph{Tracking of a basketball using multiple cameras}.
\newblock PhD thesis, University of Polytecnica de Catalunya, 2010.

\bibitem[Mondal(2014)]{Dipen2014}
Dipen~Chandra Mondal.
\newblock \emph{Multi camera soccer player tracking}.
\newblock PhD thesis, National Institute of Technology, Rourkela, India, 2014.

\bibitem[Alavi(2017)]{hosseinAlavi2017}
Amirhossein Alavi.
\newblock Investigation into tracking football players from video streams
  produced by cameras set up for {TV} broadcasting.
\newblock \emph{American Journal of Engineering Research}, 6:\penalty0 95--104,
  2017.

\bibitem[Mackowiak et~al.(2010)Mackowiak, Konieczny, Kurc, and
  Maćkowiak]{Slawomir2010}
Slawomir Mackowiak, Jacek Konieczny, Maciej Kurc, and Przemysław Maćkowiak.
\newblock Football player detection in video broadcast.
\newblock \emph{Computer Vision and Graphics, Lecture Notes in Computer
  Science}, 6375:\penalty0 118--125, 2010.

\bibitem[Cheshire et~al.(2015)Cheshire, Hu, and Chang]{Evan2015}
Evan Cheshire, Min-Chun Hu, and Ming-Hsiu Chang.
\newblock Player tracking and analysis of basketball plays.
\newblock In \emph{European Conference of Computer Vision}, 2015.

\bibitem[Ming et~al.(2009)Ming, Guodong, and Lichao]{Ming2009}
Yul Ming, Cui Guodong, and Qi~Lichao.
\newblock Player detection algorithm based on gaussian mixture models
  background modeling.
\newblock In \emph{Second International Conference on Intelligent Networks and
  Intelligent Systems}, pages 323--326, 2009.

\bibitem[Mazzeo et~al.(2008)Mazzeo, Spagnolo, Leo, and D'Orazio]{Mazzeo2008}
Pier-Luigi Mazzeo, Paolo Spagnolo, M.~Leo, and T.~D'Orazio.
\newblock Visual players detection and tracking in soccer matches.
\newblock In \emph{\uppercase{ieee} Fifth International Conference on Advanced
  Video and Signal Based Surveillance}, pages 326--333, 2008.

\bibitem[Direkoglu et~al.(2018)Direkoglu, Sah, and O'connor]{Direkoglu2018}
Cem Direkoglu, Melike Sah, and Noel O'connor.
\newblock Player detection in field sports.
\newblock \emph{Machine Vision and Applications}, 29\penalty0 (2):\penalty0
  187--206, 2018.

\bibitem[Naushad~Ali et~al.(2012)Naushad~Ali, Abdullah-Al-Wadud, and
  Lee]{Naushad2012}
MM~Naushad~Ali, Mohammad Abdullah-Al-Wadud, and Seok-Lyong Lee.
\newblock An efficient algorithm for detection of soccer ball and players.
\newblock In \emph{Signal Processing Image Processing and Pattern Recognition},
  2012.

\bibitem[Rao and Pati(2015)]{Upendra2015}
Upendra Rao and Umesh~C. Pati.
\newblock A novel algorithm for detection of soccer ball and player.
\newblock In \emph{International Conference on Communications and Signal
  Processing}, 2015.

\bibitem[Markoski et~al.(2015)Markoski, Ivankovic, Ratgeber, Predrag, and
  Glusac]{Branko2015}
Branko Markoski, Zdravko Ivankovic, Ladislav Ratgeber, Pecev Predrag, and
  Dragana Glusac.
\newblock Application of adaboost algorithm in basketball player detection.
\newblock \emph{Acta Polytechnica Hungarica}, 12:\penalty0 189--207, 2015.

\bibitem[Zhu et~al.(2006)Zhu, Xu, Huang, and Gao]{Guangyu2006}
Guangyu Zhu, Changsheng Xu, Qingming Huang, and Wen Gao.
\newblock Automatic multi-player detection and tracking in broadcast sports
  video using support vector machine and particle filter.
\newblock In \emph{\uppercase{IEEE} International Conference on Multimedia and
  Expo}, pages 1629--1632, 2006.

\bibitem[Chengjun(2018)]{Chengjun2018}
Cui Chengjun.
\newblock Player detection based on support vector machine in football videos.
\newblock \emph{International Journal of Performability Engineering},
  14\penalty0 (2):\penalty0 309--319, 2018.

\bibitem[GerkeKarsten and Schäfer(2015)]{Sebastian2015}
Sebastian GerkeKarsten and MüllerRalf Schäfer.
\newblock Soccer jersey number recognition using convolutional neural networks.
\newblock In \emph{The \uppercase{ieee} International Conference on Computer
  Vision Workshops}, pages 734--741, 2015.

\bibitem[Li et~al.(2018)Li, Xu, Liu, Li, and Wang]{Gen2018}
Gen Li, Shikun Xu, Xiang Liu, Lei Li, and Changhu Wang.
\newblock Jersey number recognition with semi-supervised spatial transformer
  network.
\newblock In \emph{\uppercase{ieee}/\uppercase{cvf} Conference on Computer
  Vision and Pattern Recognition Workshops}, pages 1864--1867, 2018.

\bibitem[Liu and Bhanu(2019)]{Hengyue2019}
Hengyue Liu and Bir Bhanu.
\newblock Pose-guided \uppercase{R-CNN} for jersey number recognition in
  sports.
\newblock In \emph{\uppercase{ieee} Conference on Computer Vision and Pattern
  Recognition Workshops}, 2019.

\bibitem[Lehuger et~al.(2007)Lehuger, Duffner, and Garcia]{Antoine2007}
Antoine Lehuger, Stefan Duffner, and Cecilia Garcia.
\newblock A robust method for automatic player detection in sport videos.
\newblock In \emph{Orange Labs, Cesson-Sévigné}, 2007.

\bibitem[Mathes and Piater(2006)]{Mathes2006}
Tom Mathes and Justus~H. Piater.
\newblock Robust non-rigid object tracking using point distribution manifolds.
\newblock \emph{Pattern Recognition, \uppercase{dagm}, Lecture Notes in
  Computer Science}, 4174:\penalty0 1--10, 2006.

\bibitem[Hayet et~al.(2005)Hayet, Mathes, Czyz, Piater, Verly, and
  Macq]{Hayet2005}
J.B Hayet, T.~Mathes, J.~Czyz, J.~Piater, J.~Verly, and B.~Macq.
\newblock A modular multi-camera framework for team sports tracking.
\newblock In \emph{\uppercase{ieee} Conference on Advanced Video and Signal
  Based Surveillance}, pages 493--498, 2005.

\bibitem[Li and Flierl(2012)]{Li2012}
H.~Li and M.~Flierl.
\newblock Sift-based multi-view cooperative tracking for soccer video.
\newblock In \emph{{IEEE} International Conference on Acoustics, Speech and
  Signal Processing}, 2012.

\bibitem[Kataoka and Aoki(2011)]{Kataoka2011}
H.~Kataoka and Y.~Aoki.
\newblock Football players and ball trajectories from single camera's image.
\newblock In \emph{17th Korea-Japan Joint Workshop on Frontiers of Computer
  Vision (\uppercase{fcv})}, 2011.

\bibitem[Manafifard et~al.(2017{\natexlab{b}})Manafifard, Ebadi, and
  Abrishami]{Manafifardd2017}
M~Manafifard, H~Ebadi, and H~Moghaddam Abrishami.
\newblock Appearance-based multiple hypothesis tracking: Application to soccer
  broadcast videos analysis.
\newblock \emph{Signal Processing: Image Communication}, 55:\penalty0 157--170,
  2017{\natexlab{b}}.

\bibitem[Petsas and Kaimakis(2016)]{Panagiotis2016}
Panagiotis Petsas and Paris Kaimakis.
\newblock Soccer player tracking using particle filters.
\newblock In \emph{\uppercase{ieee} International Symposium on Signal
  Processing and Information Technology}, pages 57--62, 2016.

\bibitem[Yang and Li(2017)]{Yang2017}
Ying Yang and Danyang Li.
\newblock Robust player detection and tracking in broadcast soccer video based
  on enhanced particle filter.
\newblock \emph{Journal of Visual Communication and Image Representation},
  46:\penalty0 81--94, 2017.

\bibitem[Dearden et~al.(2006)Dearden, Demiris, and Grau]{Anthony2006}
Anthony Dearden, Yiannis Demiris, and Oliver Grau.
\newblock Tracking football player movement from a single moving camera using
  particle filters.
\newblock In \emph{The 3rd European Conference on Visual Media Production -
  Part of the 2nd Multimedia Conference}, pages 29--37, 2006.

\bibitem[{de Pádua} et~al.(2015){de Pádua}, {Pádua}, {Sousa}, and
  d.~A.~{Pereira}]{Pedro2015}
P.~H.~C. {de Pádua}, F.~L.~C. {Pádua}, M.~T.~D. {Sousa}, and
  M.~d.~A.~{Pereira}.
\newblock Particle filter-based predictive tracking of futsal players from a
  single stationary camera.
\newblock In \emph{SIBGRAPI Conference on Graphics, Patterns and Images}, pages
  134--141, 2015.
\newblock \doi{10.1109/SIBGRAPI.2015.10}.

\bibitem[Makandar and Mulimani(2018)]{Aziz2018}
Aziz Makandar and Daneshwari Mulimani.
\newblock Analysis of multiple object detection using kalman filter in sports
  video.
\newblock In \emph{\uppercase{ijsa} Proceedings on National Conference on
  Computer Science and Information Technology}, pages 13--15, 2018.

\bibitem[Kim and Kim(2009)]{Jong2009}
Jong-Yun Kim and Tae-Yong Kim.
\newblock Soccer ball tracking using dynamic kalman filter with velocity
  control.
\newblock In \emph{Sixth International Conference on Computer Graphics, Imaging
  and Visualization}, pages 367--374, 2009.

\bibitem[Liu et~al.(2011)Liu, Liu, and Huang]{Liu2011}
Yun Liu, Xueying Liu, and Chao Huang.
\newblock A new method for shot identification in basketball video.
\newblock \emph{Journal Of Software}, 6\penalty0 (8):\penalty0 1468--1475,
  2011.

\bibitem[Hanzra and Rossi(2013)]{Bikramjot2013}
Bikramjot~Singh Hanzra and Romain Rossi.
\newblock Automatic cameraman for dynamic video acquisition of football match.
\newblock In \emph{\uppercase{ieee} Proceedings of the Second International
  Conference on Image Information Processing (\uppercase{iciip})}, pages
  142--147, 2013.

\bibitem[Beetz et~al.(2007)Beetz, Gedikli, Bandouch, Kirchlechner, v.Hoyningen
  Huene, and Perzylo]{Michael2007}
Michael Beetz, Suat Gedikli, Jan Bandouch, Bernhard Kirchlechner, Nico
  v.Hoyningen Huene, and Alexande Perzylo.
\newblock Visually tracking football games based on tv broadcasts.
\newblock \emph{International Joint Conference on Artificial
  Intelligence(\uppercase{ijcai})}, 7:\penalty0 2066--2071, 2007.

\bibitem[Patil et~al.(2018)Patil, Salve, Pawar, and MP]{Pratik2018}
Pratik Patil, Rebecca Salve, Karanjit Pawar, and Atre MP.
\newblock Offside detection in the game of football using contour mapping.
\newblock \emph{International Journal of Research in Engineering and Science
  (\uppercase{ijres})}, 6\penalty0 (4):\penalty0 66--69, 2018.

\bibitem[Lefèvre et~al.(2000)Lefèvre, Fluck, Maillard, and
  Vincent]{Sebastien2000}
Sébastien Lefèvre, Cyril Fluck, Benjamin Maillard, and Nicole Vincent.
\newblock A fast snake-based method to track football player.
\newblock In \emph{Proceedings of the \uppercase{iarp} Conference on Machine
  Vision Applications (\uppercase{iarp mva})}, 2000.

\bibitem[Lefèvre et~al.(2002)Lefèvre, Jean-Pierre, Piron, and
  Vincent]{Sebastien2002}
Sébastien Lefèvre, Gerard Jean-Pierre, Aurelie Piron, and Nicole Vincent.
\newblock An extended snake model for real-time multiple object tracking.
\newblock In \emph{Proceedings of Advanced Concepts for Intelligent Vision
  Systems (\uppercase{acivs})}, 2002.

\bibitem[Lin(2018)]{Maochun2017}
Maochun Lin.
\newblock Contour tracking algorithm for dynamic image of basketball shooting
  arm.
\newblock \emph{Journal of Discrete Mathematical Sciences and Cryptography},
  21\penalty0 (2):\penalty0 299--304, 2018.

\bibitem[Chakraborty and Meher(2013)]{Bodhisattwa2013}
Bodhisattwa Chakraborty and Sukadev Meher.
\newblock A real-time trajectory-based ball detection-and-tracking framework
  for basketball video.
\newblock \emph{Journal of Optics}, 42\penalty0 (2):\penalty0 156--170, 2013.

\bibitem[Naidoo and Tapamo(2006)]{Wayne2006}
Wayne~Chelliah Naidoo and Jules~Raymond Tapamo.
\newblock Soccer video analysis by ball, player and referee tracking.
\newblock In \emph{Proceedings of the Annual Research Conference of the South
  African Institute of Computer Scientists and Information Technologists
  (\uppercase{saicsit}) on \uppercase{it} Research in Developing Countries},
  pages 51--60, 2006.

\bibitem[Huang et~al.(2007)Huang, Llach, and Bhagavathy]{Huang2007}
Yu~Huang, Joan Llach, and Sitaram Bhagavathy.
\newblock Players and ball detection in soccer videos based on color
  segmentation and shape analysis.
\newblock In \emph{Multimedia Content Analysis and Mining, International
  Workshop, \uppercase{mcam}}, 2007.

\bibitem[Chakraborty and Meher(2012)]{Bodhisattwa2012}
Bodhisattwa Chakraborty and Sukadev Meher.
\newblock Real-time position estimation and tracking of a basketball.
\newblock In \emph{\uppercase{ieee} International Conference on Signal
  Processing, Computing and Control}, pages 1--6, 2012.

\bibitem[Figueroa et~al.(2004)Figueroa, Leite, Barros, Cohen, and
  Medioni]{Figueroa2004}
P.~Figueroa, N.~Leite, R.M.L. Barros, I.~Cohen, and G.~Medioni.
\newblock Tracking soccer players using the graph representation.
\newblock In \emph{Proceedings of the 17th International Conference on Pattern
  Recognition}, pages 787--790, 2004.

\bibitem[Pallavi et~al.(2008)Pallavi, Mukherjee, Majumdar, and
  Sural]{Pallavi2008}
V.~Pallavi, Jayant Mukherjee, Arun~K. Majumdar, and Shamik Sural.
\newblock Graph-based multiplayer detection and tracking in broadcast soccer
  videos.
\newblock \emph{\uppercase{ieee} Transactions on Multimedia}, 10\penalty0
  (5):\penalty0 794--805, 2008.

\bibitem[Xing et~al.(2011)Xing, Ai, Liu, and Lao]{Junliang20011}
Junliang Xing, Haizhou Ai, Liwei Liu, and Shihong Lao.
\newblock Multiple player tracking in sports video: A dual-mode two-way
  bayesian inference approach with progressive observation modeling.
\newblock \emph{{ieee} Transactions on Image Processing}, 20\penalty0
  (6):\penalty0 1652--1667, 2011.

\bibitem[Chen et~al.(2017)Chen, Chang, and Hsiao]{Chen2017}
Liang-Hua Chen, Hsin-Wen Chang, and Hsiang-An Hsiao.
\newblock Player trajectory reconstruction from broadcast basketball video.
\newblock In \emph{\uppercase{icbip} Proceedings of the 2nd International
  Conference on Biomedical Signal and Image Processing}, pages 72--76, 2017.

\bibitem[Maksai et~al.(2015)Maksai, Wang, and Fua]{Andrii2015}
Andrii Maksai, Xinchao Wang, and Pascal Fua.
\newblock What players do with the ball: A physically constrained interaction
  modeling.
\newblock In \emph{\uppercase{ieee} Conference on Computer Vision and Pattern
  Recognition}, pages 972--981, 2015.

\bibitem[Septier et~al.(2011)Septier, Cornebise, J.Godsill, and
  Delignon]{Septier2011}
François Septier, Julien Cornebise, Simon J.Godsill, and Yves Delignon.
\newblock A comparative study of montecarlo methods for multitarget tracking.
\newblock In \emph{\uppercase{ieee} Statistical Signal Processing Workshop},
  pages 205--208, 2011.

\bibitem[Liu et~al.(2009)Liu, Tong, and Li]{Liu2009}
Jia Liu, Xiaofeng Tong, and Wenlong Li.
\newblock Automatic player detection, labeling and tracking in broadcast soccer
  video.
\newblock \emph{Pattern Recognition Letters}, 30\penalty0 (2):\penalty0
  103--113, 2009.

\bibitem[Abbott and Williams(2009)]{Robert2009}
Robert~G. Abbott and Lance Williams.
\newblock Multiple target tracking with lazy background subtraction and
  connected components analysis.
\newblock \emph{Machine Vision and Applications}, 20\penalty0 (2):\penalty0
  93--101, 2009.

\bibitem[Kamble et~al.(2019)Kamble, Keskar, and Bhurchandi]{Kamble2019}
P.R Kamble, A.G Keskar, and K.M Bhurchandi.
\newblock A deep learning ball tracking system in soccer videos.
\newblock \emph{Opto-Electronics Review}, 27\penalty0 (1):\penalty0 58--69,
  2019.

\bibitem[Lu et~al.(2017)Lu, Chen, J.Little, and He]{KeyuLu2017}
Keyu Lu, Jianhui Chen, James J.Little, and Hangen He.
\newblock Light cascaded convolutional neural networks for accurate player
  detection.
\newblock In \emph{Proceedings of the British Machine Vision Conference
  (\uppercase{bmvc})}, pages 173.1--173.13. {bmva} Press, 2017.

\bibitem[Buric et~al.(2019)Buric, Pobar, and Ivasic-Kos]{Matija2019}
Matija Buric, Miran Pobar, and Marina Ivasic-Kos.
\newblock Adapting {yolo} network for ball and player detection.
\newblock In \emph{Proceedings of the 8th International Conference on Pattern
  Recognition Applications and Methods}, volume~1, pages 845--851, 2019.

\bibitem[Yoon et~al.(2019)Yoon, Hwang, Choi, Joo, and Oh]{Young2019}
Young Yoon, Heesu Hwang, Yongjun Choi, Minbeom Joo, and Hyeyoon Oh.
\newblock Analyzing basketball movements and pass relationships using realtime
  object tracking techniques based on deep learning.
\newblock \emph{\uppercase{ieee} Access}, 7:\penalty0 56564--56576, 2019.

\bibitem[Long(2019)]{Long2019}
Teng Long.
\newblock Research on application of athlete gesture tracking algorithms based
  on deep learning.
\newblock \emph{Journal of Ambient Intelligence and Humanized Computing},
  -:\penalty0 1--9, 2019.

\end{thebibliography}






\end{document}